\documentclass[sigconf]{acmart}

\usepackage{xspace}
\newcommand{\projectname}{\textsc{CirCNN}\xspace}
\usepackage[boxed,ruled]{algorithm2e}

\usepackage{booktabs} % For formal tables

\begin{document}
% Copyright
%\setcopyright{none}
%\setcopyright{acmcopyright}
%\setcopyright{acmlicensed}
\setcopyright{rightsretained}
%\setcopyright{usgov}
%\setcopyright{usgovmixed}
%\setcopyright{cagov}
%\setcopyright{cagovmixed}
%\copyrightyear{2017} 
%\acmYear{2017}  
%\setcopyright{acmcopyright}
%\acmConference[MICRO'17]{The 50th Annual IEEE/ACM International Symposium on Microarchitecture}{October 14-18, 2017}{Boston, MA, USA}
%\acmPrice{15.00}
%\acmDOI{10.1145/XXXXX.XXXXXX}
%\acmISBN{978-1-4503-4952-9}

\copyrightyear{2017}
\acmYear{2017}
\setcopyright{acmcopyright}
\acmConference{MICRO-50}{October 14--18, 2017}{Cambridge, MA,
USA}\acmPrice{15.00}\acmDOI{10.1145/3123939.3124552}
\acmISBN{978-1-4503-4952-9/17/10}

%\title{{\projectname}: Accelerating and Compressing Deep}
\title[{\projectname}: Accelerating and Compressing Deep Neural Networks]{{\projectname}: Accelerating and Compressing Deep Neural Networks 
Using Block-Circulant Weight Matrices}
\author{Caiwen Ding$^{+,1}$, Siyu Liao$^{+,2}$, Yanzhi Wang$^{+,1}$, Zhe Li$^{1}$, Ning Liu$^{1}$, Youwei Zhuo$^3$, Chao Wang$^3$, Xuehai Qian$^3$, Yu Bai$^4$, Geng Yuan$^{1}$, Xiaolong Ma$^{1}$, Yipeng Zhang$^{1}$, Jian Tang$^{1}$, Qinru Qiu$^{1}$, Xue Lin$^5$, Bo Yuan$^{2}$}
\affiliation{
  \institution{
  $^+$These authors contributed equally.\\
  $^1$Syracuse University, $^2$City University of New York, City College, $^3$University of Southern California, $^4$California State University Fullerton, $^5$Northeastern University }
}
\email{{cading,ywang393,zli89,nliu03,geyuan,xma27,yzhan139,jtang02,qiqiu}@syr.edu,
sliao2@gradcenter.cuny.edu, {youweizh,wang484,xuehai.qian}@usc.edu,
ybai@exchange.fullerton.edu, xue.lin@northeastern.edu, byuan@ccny.cuny.edu
}

%\author{Second Author}
%\affiliation{
%  \institution{Institution2}
%}
%\email{name2@domain2}

% List the first author for headers
\renewcommand{\shortauthors}{C. Ding et al.}

\begin{abstract}
Large-scale deep neural networks (DNNs) are both compute and memory intensive.
%making their deployments extremely challenging. 
As the size of DNNs continues to grow, it is critical to 
improve the energy efficiency and performance 
while maintaining accuracy. 
%the need to improve their
%energy efficiency and performance of deep learning systems while maintaining their quality of results has
%become an important design task. 
%To push the scalability and energy efficiency of DNNs toward their theoretical limits, 
%the fundamental improvements in the computing paradigm and associated hardware acceleration are required.
%Such advance will enable rapid up-scaling and 
%wider adoption of deep learning systems.
%to push the scalability and energy efficiency of deep learning systems towards their theoretical limits, thereby accommodating the rapid up-scaling and wide utilization of deep learning systems.
For DNNs, the model size is an important factor
affecting performance, scalability and energy efficiency.
% for deep learning systems and DNNs.
Weight pruning achieves good compression
ratios 
%for model reduction in deep learning systems, and can achieve a good reduction ratio. 
but suffers from three drawbacks:
{\em 1)} the irregular network structure after pruning, which affects 
performance and throughput;  
%of the potentially irregular network structure after pruning;
{\em 2)} the increased training complexity; and
{\em 3)} the lack of rigorous guarantee of compression ratio and inference accuracy.

%and the lack of theoretical guarantee of compression ratio and test accuracy.
%these work have limitations on the potentially irregular network structure after pruning, the increased complexity in the training process, and the lack of theoretical guarantee of compression ratio and test accuracy. 
To overcome these limitations, this paper proposes \projectname,
a principled approach to represent weights and process neural networks using 
{\em block-circulant} matrices.
% for weight representation.
%We overcome these limitations by proposing block-circulant matrices for weight representation in deep learning systems.
\projectname utilizes the {\em Fast Fourier Transform 
(FFT)}-based fast multiplication, {\em simultaneously} reducing 
the computational complexity (both in inference and training) from O($n^2$) to O($n\log n$) and 
the storage complexity from O($n^2$) to O($n$), with negligible accuracy loss. 
Compared to other approaches, \projectname is distinct due to its mathematical rigor: 
the DNNs based on \projectname can converge to the same ``effectiveness'' as DNNs without compression.
We propose the \projectname architecture, a universal DNN inference engine that can be implemented in various hardware/software 
platforms with configurable network architecture (e.g., layer type,
size, scales, etc.).
In \projectname architecture:
{\em 1)} Due to the recursive property, {\em FFT can be used as the key computing kernel}, which ensures 
universal and small-footprint implementations.
{\em 2)} The {\em compressed but regular} network structure avoids the pitfalls of the 
network pruning and facilitates high performance and throughput with
highly pipelined and parallel design.
%To demonstrate the performance and energy efficiency,
To demonstrate the performance and energy efficiency,
we test \projectname in FPGA, ASIC and embedded processors.
Our results show that \projectname architecture achieves
very high energy efficiency and 
performance with a small hardware footprint.
Based on the FPGA implementation and ASIC synthesis results, 
\projectname achieves 6 - 102X energy efficiency improvements compared with the best state-of-the-art results.

\end{abstract}

%\begin{CCSXML}
%<ccs2012>
%  <concept_id>10010520.10010553.10010562</concept_id>
%  <concept_desc>Computer systems organization~Embedded systems</concept_desc>
%  <concept_significance>500</concept_significance>
%  <concept_id>10010520.10010575.10010755</concept_id>
%  <concept_desc>Computer systems organization~Redundancy</concept_desc>
%  <concept_significance>300</concept_significance>
%  <concept_id>10010520.10010553.10010554</concept_id>
%  <concept_desc>Computer systems organization~Robotics</concept_desc>
%  <concept_significance>100</concept_significance>
%  <concept_id>10003033.10003083.10003095</concept_id>
%  <concept_desc>Networks~Network reliability</concept_desc>
%  <concept_significance>100</concept_significance>
%</ccs2012>  
%\end{CCSXML}
%
%\ccsdesc[500]{Computer systems organization~Embedded systems}




\maketitle

\section{Introduction}
From the end of the first decade of the 21st century, 
%Starting from the late 2000's, 
neural networks have been experiencing a phenomenal resurgence thanks to the big data and the significant advances in processing speeds. Large-scale deep neural networks (DNNs) have been able to deliver impressive results in many challenging problems. 
For instance, DNNs have led to breakthroughs in object recognition accuracy on the ImageNet dataset \cite{deng2009imagenet}, even achieving human-level performance for face recognition \cite{taigman2014deepface}. Such promising results triggered the revolution of several traditional and emerging real-world applications, such as self-driving systems \cite{huval2015empirical}, automatic machine translations \cite{collobert2008unified}, drug discovery and toxicology \cite{burbidge2001drug}. As a result,
both academia and industry show the rising interests
with significant resources devoted to 
investigation, improvement, and promotion of deep learning 
methods and systems.
%, both of which have devoted significant resources to investigate, improve, and promote explorations of deep learning methods and systems.

One of the key enablers of the unprecedented success of deep learning is the availability of very large models. 
Modern DNNs typically consist of multiple cascaded layers, and at least millions to hundreds of millions of parameters (i.e., weights) for the entire model \cite{krizhevsky2012imagenet,karpathy2015deep,catanzaro2013deep,simonyan2014very}. The larger-scale neural networks tend to 
enable the extraction of more complex high-level features, and  therefore,
lead to a significant improvement of the overall accuracy \cite{le2013building,ciregan2012multi,schmidhuber2015deep}. 
On the other side, the layered deep structure and 
large model sizes also demand increasing computational capability and memory requirements. In order to achieve higher scalability, performance, and energy efficiency for deep learning systems, two orthogonal research and development trends have both attracted enormous interests.

%Despite the advantage of improved overall accuracy, the deep layered structure and large model sizes increase the computational complexity and memory requirements. 
The first trend is the {\em hardware acceleration of DNNs}, 
which has been extensively investigated in both industry and academia. As a representative technique, \textbf{FPGA-based} accelerators offer
%can offer the advantages of 
good programmability, high degree of parallelism and short development cycle. 
%Important progresses have been reported on FPGA accelerations of original DNNs
FPGA has been used to accelerate the original DNNs \cite{suda2016throughput,qiu2016going,zhang2016caffeine,zhang2016energy,mahajan2016tabla}, binary neural networks \cite{zhao2017accelerating,umuroglu2017finn}, and more recently,  DNNs with model compression \cite{han2017ese}. Alternatively, \textbf{ASIC-based} implementations 
%of deep learning systems 
have been recently explored to overcome the limitations of general-purpose computing approaches. A number of major high-tech companies have announced their ASIC chip designs of the DNN inference framework, such as Intel, Google, etc. \cite{company1,company2}. In academia, three representative works at the architectural level are Eyeriss \cite{chen2017eyeriss}, EIE \cite{han2016eie}, and the DianNao family \cite{chen2014diannao,chen2014dadiannao,du2015shidiannao}, which focus specifically on the convolutional layers, the fully-connected layers, and the memory design/organization, respectively. 
There are a number of recent tapeouts of hardware deep learning systems
\cite{chen2017eyeriss,reagen2016minerva,desoli201714,moons201714,sim20161,whatmough201714,bang201714}.

These prior works mainly focus on the inference phase of DNNs, and
usually suffer from the frequent accesses to off-chip DRAM systems 
(e.g., when large-scale DNNs are used for ImageNet dataset). 
This is because the limited on-chip SRAM memory can hardly accommodate large model sizes. 
Unfortunately, off-chip DRAM accesses consume significant energy. 
%Accessing off-chip DRAM is highly energy inefficient. 
The recent studies \cite{han2015learning,han2015deep} show that
the per-bit access energy of off-chip DRAM memory is 200$\times$ compared with on-chip SRAM.
Therefore, it can easily dominate the whole system power consumption.
%Besides, it is also desirable to achieve algorithmic-level accelerations to accommodate the further DNN scaling, instead of simply adding more and more hardware devices.

The energy efficiency challenge of large models motivates the 
second trend: {\em model compression}.
Several algorithm-level techniques have been proposed 
to compress models and accelerate DNNs, including weight quantization \cite{wu2016mobile,lin2016fixed}, connection pruning \cite{han2015learning,han2015deep}, and low rank approximation \cite{jaderberg2014speeding, tai2015convolutional}. These approaches can offer a reasonable parameter reduction (e.g., by $9\times$ to $13\times$ in \cite{han2015learning,han2015deep}) with minor accuracy degradation. However, they suffer from the three drawbacks: 
{\em 1)} the sparsity regularization and pruning typically 
result in an irregular network structure, thereby undermining the compression ratio and limiting performance and throughput \cite{yu2017scalpel}; 
{\em 2)} the training complexity is increased due to the
additional pruning process \cite{han2015learning,han2015deep} or 
low rank approximation step \cite{jaderberg2014speeding,tai2015convolutional}, etc.; 
{\em 3)} the compression ratios depending on network are heuristic and cannot be precisely controlled.
%, not to mention a mathematically rigorous proof of the effectiveness.

We believe that an {\em ideal} model compression technique should:
{\em i)} maintain {\em regular} network structure;
{\em ii)} reduce the complexity for 
{\em both inference and training}, and, most importantly, 
{\em iii)} retain a {\em rigorous} mathematical fundation on compression ratio and accuracy.

\begin{figure}[t]
\begin{center}

\includegraphics[width = 0.46\textwidth]{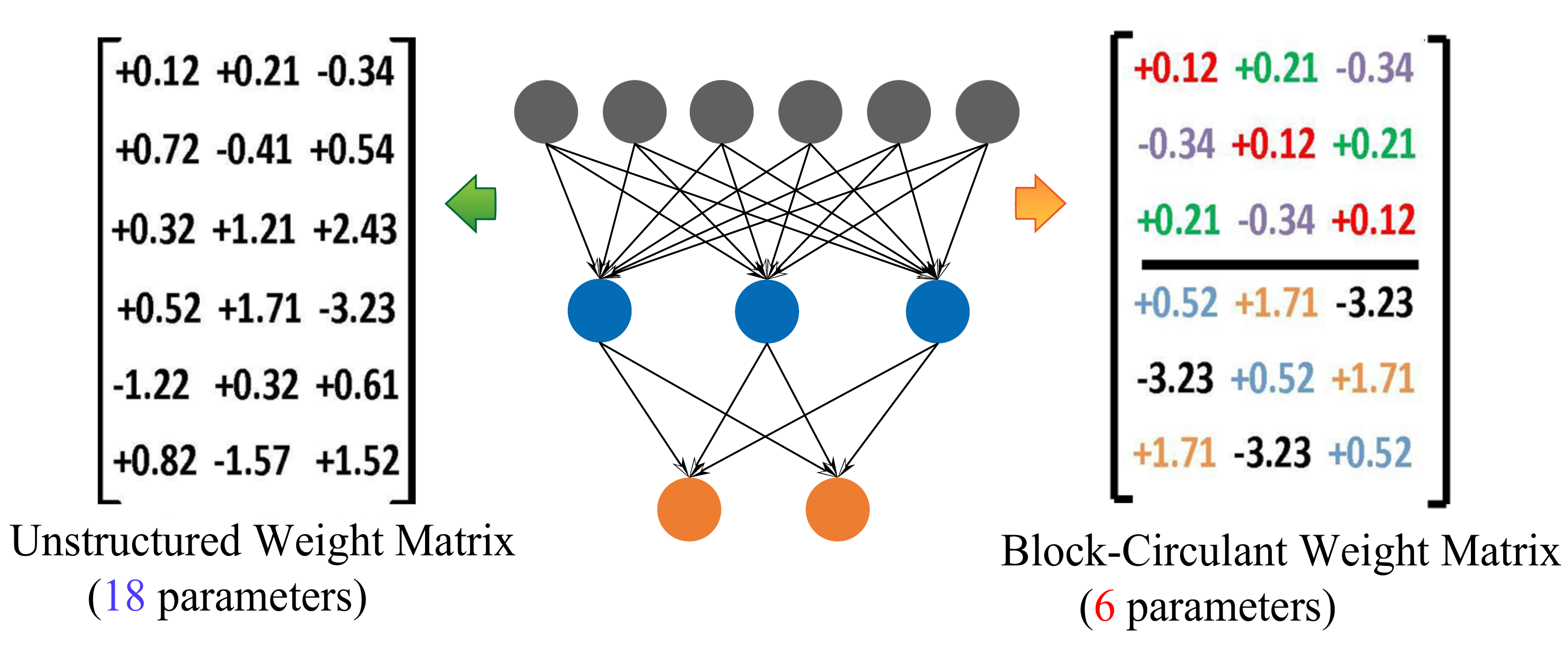}
\vspace{-0.6cm}
\caption{Block-circulant Matrices for weight representation.}
\label{fig_BlockCirculant}
\end{center}
\vspace{-0.6cm}
\end{figure}

As an effort to achieve the three goals, 
we propose \projectname,
a principled approach to represent weights and process neural networks using 
{\em block-circulant} matrices \cite{pan2012structured}.
The concept of the block-circulant matrix compared to the 
ordinary unstructured matrix is shown in Fig. \ref{fig_BlockCirculant}.
In a square circulant matrix, 
each row (or column) vector is the circulant reformat of the other row (column) vectors.
A non-squared matrix could be represented by a set of 
square circulant submatrices (blocks).
Therefore, by representing a matrix with a vector, 
the first benefit of \projectname is {\em storage size reduction}.
In Fig. \ref{fig_BlockCirculant}, 
the unstructured $6 \times 3$ weight matrix (on the left) holds 18 parameters. 
Suppose we can represent the weights using two $3 \times 3$ circulant 
matrices (on the right), we just need to store {\em 6 parameters}, 
easily leading to 3x model size reduction.
Intuitively, the reduction ratio is determined by the block size 
of the circulant submatrices: larger block size leads
to high compression ratio.
In general, the {\bf storage complexity is
reduced from O($n^2$) to O($n$)}.

The second benefit of \projectname is 
{\em computational complexity reduction}.
We explain the insights using a fully-connected layer of DNN,
which can be represented as 
$\bf{y}=\psi(\bf{W}\bf{x}+\mathbf{\theta})$, where vectors $\bf{x}$ and $\bf{y}$ represent the outputs of all neurons in the previous layer and the current layer, respectively; $\bf{W}$ is the $m$-by-$n$ weight matrix; and $\psi(\cdot)$ is activation function.
When $\bf{W}$ is a block-circulant matrix, the {\em Fast Fourier Transform (FFT)}-based fast multiplication method can be utilized, and the {\bf computational complexity is reduced from O($n^2$) to O($n\log n$)}.

It is important to understand that \projectname incurs {\em no 
conversion} between the unstructured weight matrices and 
block-circulant matrices. Instead, we {\em assume} that the 
layers can be represented by block-circulant matrices and 
the training generates a vector for each circulant 
submatrix. 
The fundamental difference is that: the current approaches
apply various compression techniques (e.g., pruning)
on the unstructured weight matrices and then retrain the network;
while \projectname directly trains the network assuming 
block-circulant structure. 
This leads to two advantages.
First, the prior work can only reduce the model size by a
heuristic factor, depending on the network, while 
\projectname provides the {\em adjustable but fixed}
reduction ratio.
Second, with the same FFT-based
fast multiplication, the computational complexity of training
is also reduced from O($n^2$) to O($n\log n$).
Unfortunately, the prior work does not reduce (or even increase)
training complexity.

Due to the storage and computational complexity reduction,
\projectname is clearly attractive. The only question is:
can a network really be represented by block-circulant matrices
with no (or negligible) accuracy loss? This question is natural,
because with the much less weights in the vectors, the network
may not be able to approximate the function of the network
with unstructured weight matrices.
Fortunately, the answer to the question is {\em YES}.
\projectname is {\em mathematically rigorous}:   
we have developed a theoretical foundation and formal proof showing that the DNNs represented by block-circulant matrices 
can converge to the same ``effectiveness" as DNNs without compression, fundamentally distinguishing our method from prior arts.
The outline of the proof is discussed in Section~\ref{proof} and 
the details are provided in technical reports \cite{proof_simple,proof}. 
 
Based on block-circulant matrix-based algorithms,
we propose \projectname architecture, --- a universal DNN inference engine that can be implemented in various hardware/software platforms with configurable network architecture (e.g., layer type, size, scales, etc.). 
Applying \projectname to neural network accelerators enables notable architectural innovations. 
{\em 1)} Due to its recursive property and its intrinsic role in 
\projectname, FFT is implemented as the {\em basic computing block}.
It ensures universal and small-footprint implementations.
{\em 2)} {\em Pipelining and parallelism} optimizations. 
Taking advantage of the compressed but regular network structures, 
we aggressively apply inter-level and intra-level pipelining
in the basic computing block.
Moreover, we can conduct joint-optimizations considering 
parallelization degree, performance and power consumption.  
{\em 3)} {\em Platform-specific optimizations} focusing on 
weight storage and memory management.

To demonstrate the performance and energy efficiency,
we test \projectname architecture in
{\em three platforms}: FPGA, ASIC and embedded processors.
Our results show that \projectname architecture achieves
very high energy efficiency and 
performance with a small hardware footprint.
Based on the FPGA implementation and ASIC synthesis results, 
\projectname achieves 6 - 102X energy efficiency improvements compared with the best state-of-the-art results.

\section{Background and Motivation}

\subsection{Deep Neural Networks}

\begin{figure}[htbp]
\begin{center}
\includegraphics[width = 0.46\textwidth]{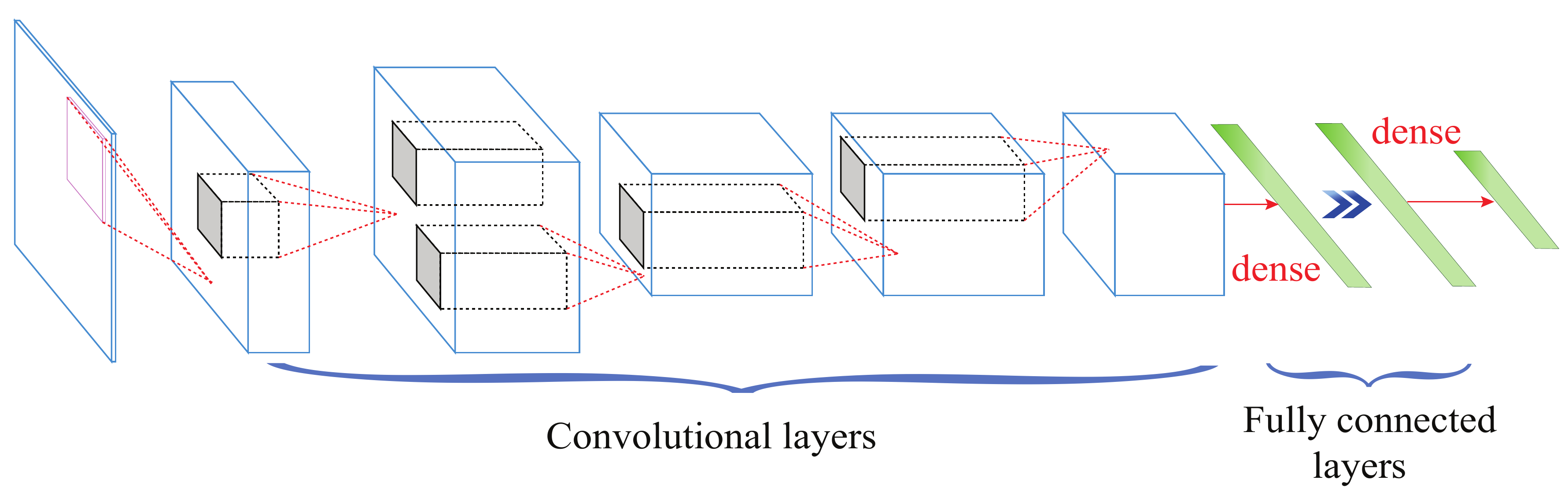}
\caption{Multi-layer structure of an example DCNN.}
\label{fig_DCNN}
\end{center}
\vspace{-0.3em}
\end{figure}

Deep learning systems can be constructed using different types of architectures, including deep convolutional neural networks (DCNNs), deep belief networks (DBNs), and recurrent neural networks (RNNs). Despite the differences in network structures and target applications, they share the {\em same construction principle}: 
multiple functional layers are cascaded together to extract features 
at multiple levels of abstraction \cite{lee2009convolutional,karpathy2014large,yu2011deep}. Fig. \ref{fig_DCNN} illustrates the multi-layer structure of an example DCNN, which consists of a stack of \emph{fully-connected layers}, \emph{convolutional layers}, and \emph{pooling layers}. These three types of layers are fundamental in deep learning systems.

\emph{\textbf{The fully-connected (FC) layer}} is the most storage-intensive layer in DNN architectures \cite{qiu2016going,zhang2016caffeine} since its neurons are fully connected with neurons in the previous layer. The computation procedure of a FC layer consists of matrix-vector arithmetics (multiplications and additions) and transformation by the activation function, as described as follows:
\begin{equation}
\bf{y}=\psi(\bf{W}\bf{x}+\mathbf{\theta})
\end{equation}
where $\textbf{W}\in \mathbb{R}^{m\times n}$ is the weight matrix of the synapses between this FC layer (with $m$ neurons) and its previous layer (with $n$ neurons); $\mathbf{\theta}\in \mathbb{R}^{m}$ is the bias vector; and $\psi(\cdot)$ is the activation function. The Rectified Linear Unit (ReLU) $\psi(x)=\max(0,x)$ is the most widely utilized in DNNs.

\emph{\textbf{The convolutional (CONV) layer}}, as the name implies, performs a two-dimensional convolution to extract features from its inputs that will be fed into subsequent layers for extracting higher-level features. A CONV layer is associated with a set of learnable filters (or kernels) \cite{lecun1998gradient}, which are activated when specific types of features are found at some spatial positions in inputs. A filter-sized moving window is applied to the inputs to obtain a set of feature maps, calculating the convolution of the filter and inputs in the moving window. Each \emph{convolutional neuron}, representing one pixel in a feature map, takes a set of inputs and the corresponding filter weights to calculate the inner-product. Given input feature map $\textbf{X}$ and the $r\times r$-sized filter (i.e., the \emph{convolutional kernel}) $\textbf{F}$, the output feature map $\textbf{Y}$ is calculated as
\begin{equation}
y_{a,b}=\sum_{i=1}^r\sum_{j=1}^r x_{a+i-1,b+j-1}\times f_{i,j},
\end{equation}
where $y_{a,b}$, $x_{a+i-1,b+j-1}$, and $f_{i,j}$ are elements in $\textbf{Y}$, $\textbf{X}$, and $\textbf{F}$, respectively. Multiple convolutional kernels can be adopted to extract different features in the same input feature map. Multiple input feature maps can be convolved with the same filter and results are summed up to derive a single feature map.

\emph{\textbf{The pooling (POOL) layer}} performs a subsampling operation on the extracted features to reduce the data dimensions and mitigate overfitting issues. Here, the subsampling operation on the inputs of pooling layer can be realized by various non-linear operations, such as max, average or L2-norm calculation. Among them, the max pooling is the dominant type of pooling strategy in state-of-the-art DCNNs due to the higher overall accuracy and convergence speed \cite{han2017ese,chen2017eyeriss}.

Among these three types of layers, the majority of computation occurs in CONV and FC layers, while the POOL layer has a relatively lower computational complexity of O($n$). The storage requirement of DNNs is due to the weight matrices $\textbf{W}$'s in the FC layers and the convolutional kernels $\textbf{F}$'s in CONV layers. As a result, the FC and CONV layers become the major research focuses on energy-efficient implementation and weight reduction of DNNs.

\subsection{DNN Weight Storage Reduction and Acceleration}

\begin{figure}[htbp]
\begin{center}
\includegraphics[width = 0.45\textwidth]{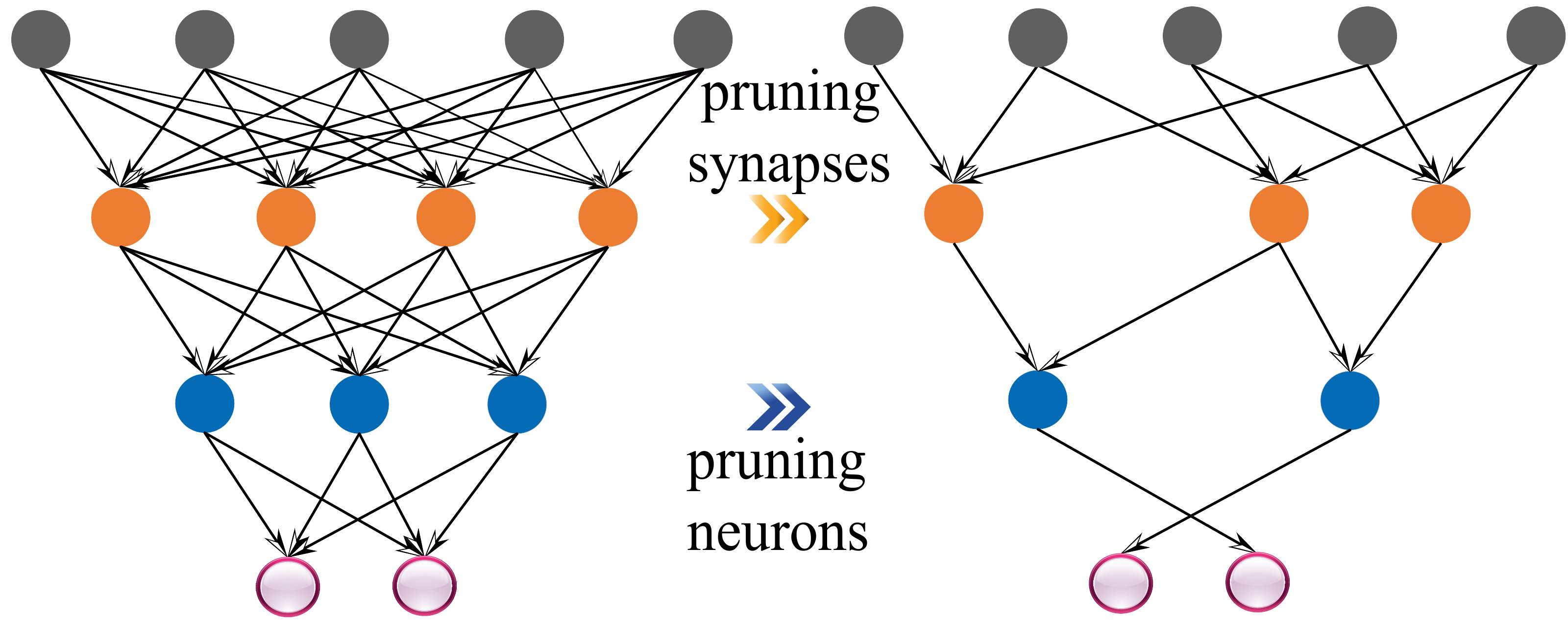}
\caption{Illustration of the heuristic weight pruning methods.}
\label{fig_Heuristics}
\end{center}
\vspace{-0.3em}
\end{figure}

Mathematical investigations have demonstrated significant sparsity and margin for weight reduction in DNNs, 
a number of prior works leverage this property to 
reduce weight storage. 
The techniques can be classified into two categories.
%The current works on weight reduction can be mainly classified into two types. 
{\em 1)} {\em Systematic} methods~\cite{xue2013restructuring,liu2015eda,chung2016simplifying} such as Singular Value Decomposition (SVD). Despite being systematic, 
these methods typically exhibit a relatively high degradation in the overall accuracy (by 5\%-10\% at 10$\times$ compression). 
{\em 2)} {\em Heuristic pruning} methods \cite{han2015deep,han2015learning,hwang2014fixed} use heuristic weight  together with weight quantization. 
These method could achieve a better 
parameter reductions, i.e., 9$\times$-13$\times$ \cite{han2015deep,han2015learning}, and a very small accuracy degradation.
However, the network structure and weight storage after pruning become {\em highly irregular} (c.f. Fig. \ref{fig_Heuristics}) and therefore indexing is always needed, which undermines the compression ratio and more importantly, the performance improvement.

Besides the pros and cons of the two approaches, 
the prior works share the following common limitations: 
{\em 1)} mainly focusing on weight reduction rather than computational complexity reduction; 
{\em 2)} only reducing the model size by a heuristic factor instead of reducing the Big-O complexity; and 
{\em 3)} performing weight pruning or applying matrix transformations based on a trained DNN model, thereby adding complexity to the training process. The third item is crucial because it may {\em limit
the scalability of future larger-scale deep learning systems}.

\subsection{FFT-Based Methods}

LeCun {\em et al.} has proposed using FFTs to accelerate the computations in the CONV layers, which applies only to a single filter in the CONV layer \cite{mathieu2013fast}.
It uses FFT to calculate the traditional inner products of filters and input feature maps, and can achieve speedup for large filter sizes (which is less common in state-of-the-art DCNNs \cite{he2016deep}). The underlying neural network structure and parameters remain {\em unchanged}. The speedup is due to filter reuse and it cannot achieve 
either asymptotic speedup in big-O notation or weight compressions (in fact additional storage space is needed).

The work most closely 
related to \projectname is \cite{cheng2015exploration}. 
It proposed to use circulant matrix
in the inference and training algorithms. 
However, it has a number of limitations. 
First, it only applied to FC layers, but not CONV layer. 
It limits the potential gain in weight reduction and performance.
Second, it uses a {\em single} circulant matrix to 
represent the weights in the whole FC layer. 
Since the number of input and output neurons are usually not 
the same, this method leads to the storage waste due to the 
padded zeros (to make the circulant matrix squared).

\subsection{Novelty of {\projectname}}

Compared with LeCun {\em et al.} \cite{mathieu2013fast},
\projectname is fundamentally different as it achieves asymptotic speedup in big-O notation and weight compression simultaneously.
Compared with \cite{cheng2015exploration},
\projectname generalizes in three significant and novel aspects. 

{\bf Supporting both FC and CONV layers}.
Unlike FC layers, the matrices in CONV layers are small filters 
(e.g., $3 \times 3$).
Instead of representing each filter as a circulant matrix, 
\projectname exploits the {\em inter-filter sparsity} among different filters. In another word, \projectname represents a matrix of 
filters, where input and output channels are the two dimensions, by 
a vector of filters. 
The support for CONV layers allow \projectname to be applied
in the whole network.

{\bf Block-circulant matrices}.
To mitigate the inefficiency due to the single large
circulant matrix used in \cite{cheng2015exploration},
\projectname uses block-circulant matrices for weight representation.
The benefits are two-fold.
First, it avoids the wasted storage/computation due to 
zero padding when the numbers of inputs and outputs are not equal.
Second, it allows us to 
derive a {\em fine-grained tradeoff between accuracy and compression/acceleration}.
Specifically, to achieve better compression ratio,
larger block size should be used, however, it may lead
to more accuracy degradation.
The smaller block sizes provide better accuracy, but less 
compression. There is no compression if the block size is 1.

{\bf Mathematical rigorousness}. 
Importantly, we perform theoretical analysis to prove that the ``effectiveness" of block-circulant matrix-based DNNs will (asymptotically) approach that of original networks without compression.
The theoretical proof also distinguishes the proposed method 
with prior work. 
The outline of the proof is discussed in Section~\ref{proof} and 
the details are provided in reports \cite{proof_simple,proof}.

\begin{figure}[t]
\begin{center}
\includegraphics[width = 0.45\textwidth]{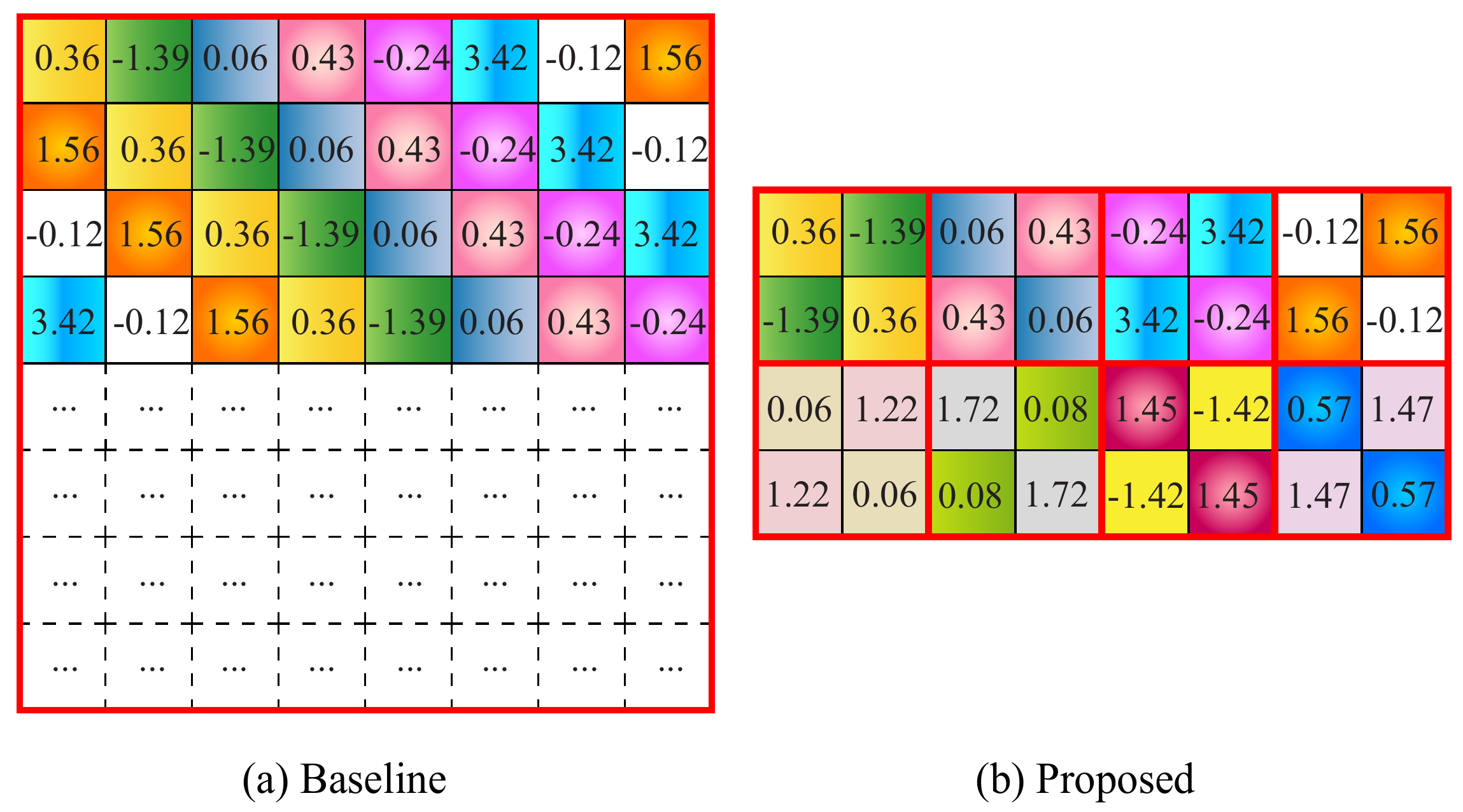}
\caption{Baseline~\cite{cheng2015exploration} and \projectname.
The baseline method (a) formulates a large, square circulant matrix for FC layer weight representation when the numbers of inputs and outputs are not equal, whereas the proposed method (b) uses the block-circulant matrix to achieve a fine-grained tradeoff of accuracy and compression/acceleration.}
\label{fig_baseline_proposed}
\end{center}
\vspace{-0.5cm}
\end{figure}

Fig.~\ref{fig_baseline_proposed} illustrates the difference 
between the baseline~\cite{cheng2015exploration} and 
\projectname.
The baseline method (a) formulates a large, square circulant matrix 
by zero padding 
for FC layer weight representation when the numbers of inputs and outputs are not equal.
In contrast, \projectname (b) uses the block-circulant matrix to
avoid storage waste and achieve a fine-grained tradeoff of accuracy and compression/acceleration.

Overall, with the novel techniques of \projectname, at algorithm level,
it is possible to achieve the {\em simultaneous} and significant reduction of {\em both computational and storage} complexity, for {\em both inference and training}. 

\section{{\projectname}: Algorithms and Foundation}
\label{algo}

%not linked for double-blind reviews.
\subsection{FC Layer Algorithm}

\begin{figure}[htbp]
\begin{center}
\includegraphics[width = 0.45\textwidth]{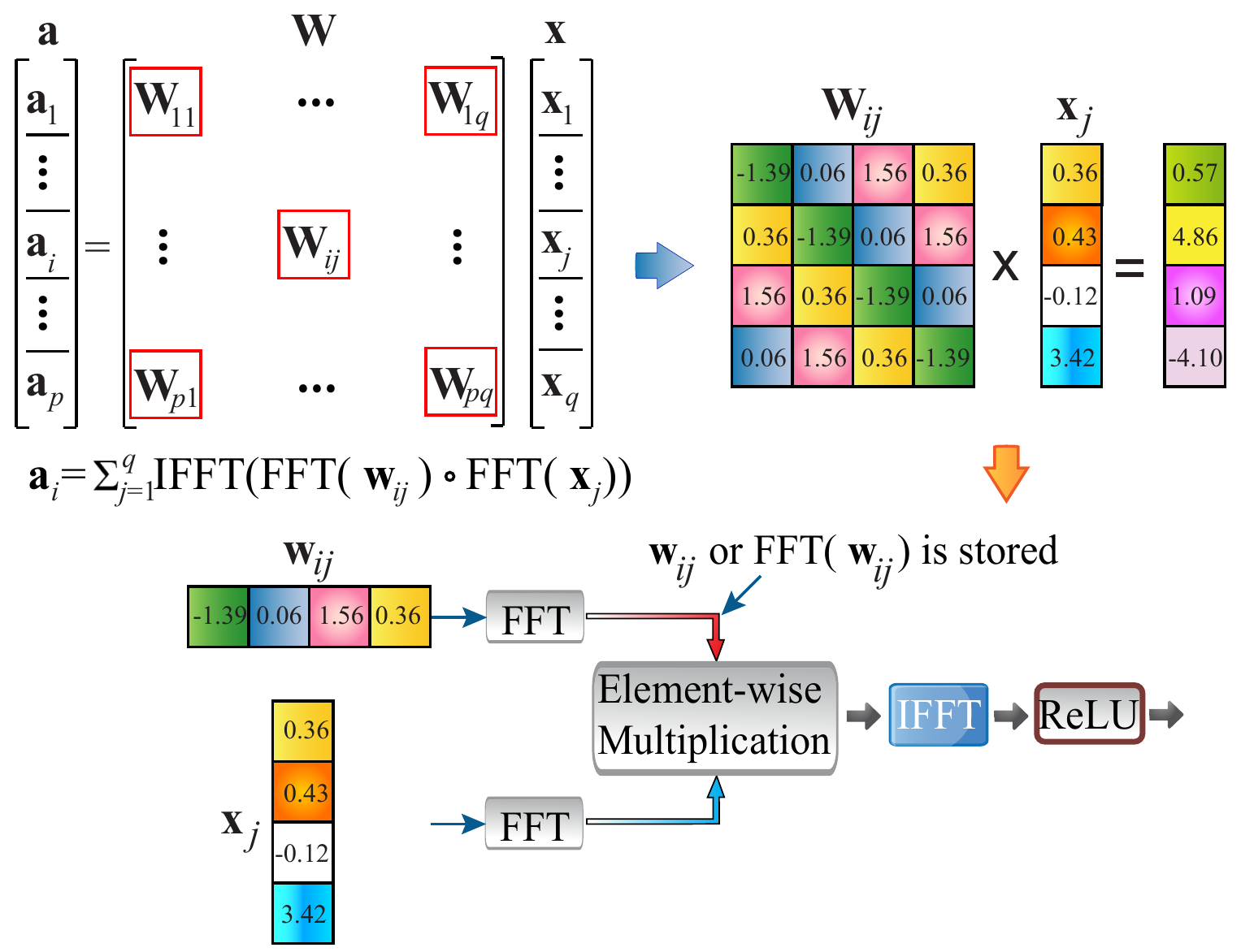}
\caption{Illustration of the calculation of $\bf{Wx}$ in the inference process of FC layer.}
\label{fig_Block_matrix}
\end{center}
\vspace{-0.3em}
\end{figure}

The key idea of block-circulant matrix-based FC layers is to partition the original arbitrary-size weight matrix $\textbf{W}\in \mathbb{R}^{m\times n}$ into 2D blocks of square sub-matrices, and each sub-matrix is a circulant matrix. 
The insights are shown in Fig. ~\ref{fig_Block_matrix}. 
Let $k$ denote the \emph{block size} (size of each sub-matrix) and 
assume there are $p \times q$ blocks after partitioning $\mathbf{W}$, where $p = m\div k$ and $q=n \div k$. Then $\mathbf{W} = [\mathbf{W}_{ij}]$, $i \in \{1 \dots p\}$, $j \in \{1 \dots q\}$. Correspondingly, the input $\mathbf{x}$ is also partitioned as $\mathbf{x} = [\mathbf{x}^T_1, \mathbf{x}^T_2, \dots, \mathbf{x}^T_q]^T$. Then, the forward propagation process in the inference phase is given by (with bias and ReLU omitted):
\begin{equation}
\mathbf{a}
=
\mathbf{Wx} 
=
\begin{bmatrix}
         \sum_{j=1}^q \mathbf{W}_{1j} \mathbf{x}_j   \\
         \sum_{j=1}^q \mathbf{W}_{2j} \mathbf{x}_j   \\
         \dots \\
         \sum_{j=1}^q \mathbf{W}_{pj} \mathbf{x}_j  
\end{bmatrix}
=
\begin{bmatrix}
         \mathbf{a}_1   \\
         \mathbf{a}_2   \\
         \dots \\
         \mathbf{a}_p
\end{bmatrix},
\end{equation}
where $\mathbf{a}_i \in \mathbb{R}^{k}$ is a column vector. Assume each circulant matrix $\mathbf{W}_{ij}$ is defined by a vector $\mathbf{w}_{ij}$, i.e., $\mathbf{w}_{ij}$ is the first row vector of $\mathbf{W}_{ij}$. Then according to the \emph{circulant convolution theorem} \cite{pan2012structured,bini1996polynomial}, the calculation of $\mathbf{W}_{ij} \mathbf{x}_j$ can be performed as $\text{IFFT}\big(\text{FFT}(\mathbf{w}_{ij})\circ\text{FFT}(\mathbf{x}_j)\big)$, where $\circ$ denotes element-wise multiplications. The operation procedure is shown on the right of
Fig. ~\ref{fig_Block_matrix}. For the inference phase, the computational complexity of this FC layer will be $O(pqk\log k)$, which is equivalent to $O(n\log n)$ for small $p$, $q$ values. Similarly, the storage complexity will be $O(pqk)$ because we only need to store $\mathbf{w}_{ij}$ or $\text{FFT}(\mathbf{w}_{ij})$ for each sub-matrix, which is equivalent to $O(n)$ for small $p$, $q$ values. 
Therefore, the simultaneous acceleration and model compression compared with the original DNN can be achieved for the inference process. Algorithm 1 illustrates the calculation of $\bf{Wx}$ in the inference process in the FC layer
of \projectname.

\SetAlgoNoLine
\begin{algorithm}[!tb]
 \KwIn{$\mathbf{w}_{ij}$'s, $\mathbf{x}$, $p$, $q$, $k$}
 \KwOut{$\mathbf{a}$}
 Initialize $\mathbf{a}$ with zeros. \\
 \For{$i \gets 1$ until p}{
 	\For{$j \gets 1$ until q} {
        $\mathbf{a}_{i}$ $\gets $ $\mathbf{a}_{i}$ + $\text{IFFT}(\text{FFT}(\mathbf{w}_{ij}) \circ \text{FFT}(\mathbf{x}_j))$\\
 	}
 }
 \Return{$\mathbf{a}$}\\
 \caption{Forward propagation process in the FC layer of \projectname}
\end{algorithm}

Next, we consider the backward propagation process in the training phase. Let $a_{il}$ be the $l$-th output element in $\mathbf{a}_i$, and $L$ denote the loss function. Then by using the chain  rule we can derive the backward propagation process as follows:
\begin{equation}
\frac{\partial L}{\partial \mathbf{w}_{ij}} 
= \sum_{l=1}^k \frac{\partial L}{\partial a_{il}} \frac{\partial a_{il}}{\partial \mathbf{w}_{ij}}
= \frac{\partial L}{\partial \mathbf{a}_i} \frac{\partial \mathbf{a}_i}{\partial \mathbf{w}_{ij}},
\end{equation}
\begin{equation}
\frac{\partial L}{\partial \mathbf{x}_{j}} 
= \sum_{i=1}^p\sum_{l=1}^k \frac{\partial L}{\partial a_{il}} \frac{\partial a_{il}}{\partial \mathbf{x}_{j}}
= \sum_{i=1}^p \frac{\partial L}{\partial \mathbf{a}_i} \frac{\partial \mathbf{a}_i}{\partial \mathbf{x}_j}.
\end{equation}
We have proved that $\frac{\partial \mathbf{a}_i}{\partial \mathbf{w}_{ij}}$ and $\frac{\partial\mathbf{a}_i}{\partial \mathbf{x}_j}$ are block-circulant matrices. Therefore, $\frac{\partial L}{\partial \mathbf{w}_{ij}}$ and $\frac{\partial L}{\partial \mathbf{a}_i} \frac{\partial \mathbf{a}_i}{\partial \mathbf{x}_j}$ can be calculated as the ``FFT$\rightarrow$element-wise multiplication$\rightarrow$IFFT'' procedure and is equivalent to $O(n\log n)$ computational complexity per layer. Algorithm 2 illustrates
backward propagation process in the FC layer
of \projectname.

\SetAlgoNoLine
\begin{algorithm}[!tb]
 \KwIn{$\frac{\partial L}{\partial \mathbf{a}}$, $\mathbf{w}_{ij}$'s, $\mathbf{x}$, $p$, $q$, $k$}
 \KwOut{$\frac{\partial L}{\partial \mathbf{w}_{ij}}$'s, $\frac{\partial L}{\partial \mathbf{x}}$}
 Initialize $\frac{\partial L}{\partial \mathbf{w}_{ij}}$'s and $\frac{\partial L}{\partial \mathbf{x}}$ with zeros.\\
 \For{$i \gets 1$ until p}{
 	\For{$j \gets 1$ until q}{
        $\frac{\partial L}{\partial \mathbf{w}_{ij}} \gets \text{IFFT}(\text{FFT}(\frac{\partial L}{\partial \mathbf{a}_i})\circ \text{FFT}(\mathbf{x'}_j) )$\\
        $\frac{\partial L}{\partial \mathbf{x}_j} \gets \frac{\partial L}{\partial \mathbf{x}_j} + \text{IFFT}( \text{FFT}(\frac{\partial L}{\partial \mathbf{a}_i})\circ\text{FFT}(\mathbf{w}_{ij}))$\\
 	}
 }
 \Return{$\frac{\partial L}{\partial \mathbf{w}_{ij}}$'s, $\frac{\partial L}{\partial \mathbf{x}}$}\\
 \caption{Backward propagation process in the FC layer of \projectname}
\end{algorithm}

In \projectname, the inference and training constitute an
integrated framework where the reduction of computational complexity
can be gained for both. 
We directly train the vectors $\mathbf{w}_{ij}$'s, corresponding to the circulant sub-matrices $\mathbf{W}_{ij}$'s, in each layer using Algorithm 2. Clearly, the network after such training procedure
naturally follows the block-circulant matrix structure. 
It is a key advantage of \projectname compared with prior
works which require additional steps on a trained neural network.

\subsection{CONV Layer Algorithm}

%The use of block-circulant matrices can also enable significant reduction in computational and storage complexities of the CONV layer. 
In practical DNN models, the CONV layers are often associated with multiple input and multiple output feature maps. As a result, the computation in the CONV layer can be expressed in the format of tensor computations as below:
\begin{equation}
\mathcal{Y}(x,y,p)=\sum_{i=1}^r\sum_{j=1}^r\sum_{c=1}^C\mathcal{F}(i,j,c,p)\mathcal{X}(x+i-1,y+j-1,c),
\end{equation}
where $\mathcal{X}\in\mathbb{R}^{W\times H\times C}$, $\mathcal{Y}\in\mathbb{R}^{(W-r+1)\times(H-r+1)\times P}$, $\mathcal{F}\in\mathbb{R}^{r\times r\times C\times P}$ represent the input, output, and weight ``tensors" of the CONV layer, respectively. Here, $W$ and $H$ are the spatial dimensions of the input maps, $C$ is the number of input maps, $r$ is the size of the convolutional kernel, and $P$ is the number of output maps.

We generalize the concept of ``block-circulant structure" to the rank-4 tensor ($\mathcal{F}$) in the CONV layer, i.e., \emph{all the slices of the form $\mathcal{F}(\cdot,\cdot,i,j)$ are circulant matrices.} Next, we reformulate the inference and training algorithms of the CONV layer to matrix operations. We use the inference process as an example, and the training process can be formulated in a similar way.

\begin{figure}[htbp]
\begin{center}
\includegraphics[width = 0.35\textwidth]{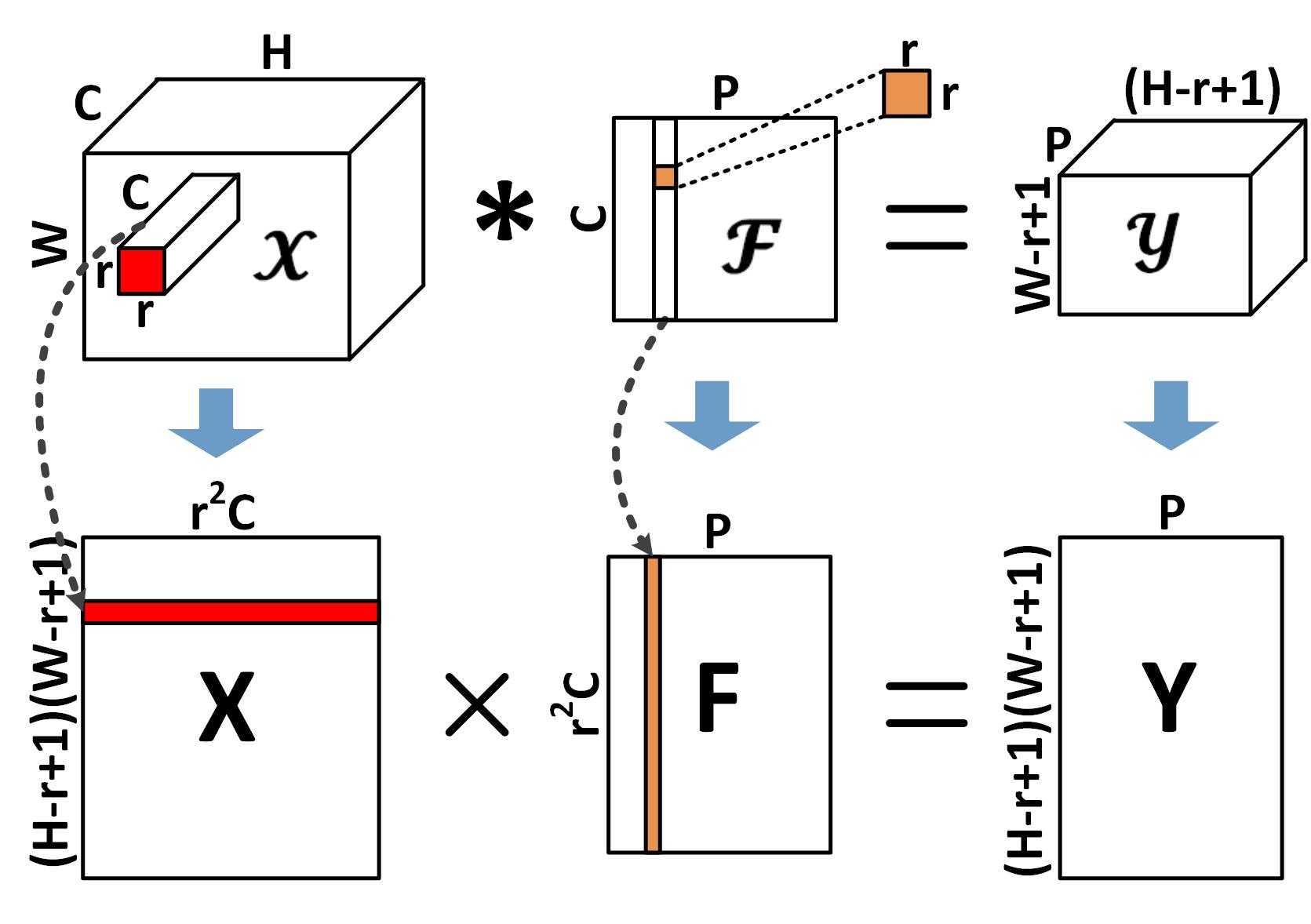}
\caption{Reformulation of Eqn. (6) to matrix multiplication.}
\label{fig_reformulation}
\end{center}
\vspace{-0.3em}
\end{figure}

Software tools such as Caffe provide an efficient methodology of transforming tensor-based operations in the CONV layer to matrix-based operations \cite{jia2014caffe,vedaldi2015matconvnet}, in order to enhance the implementation efficiency (GPUs are optimized for matrix operations.) Fig.~\ref{fig_reformulation} illustrates the application of the method to reformulate Eqn. (6) to the matrix multiplication $\mathbf{Y}=\mathbf{XF}$, where $\mathbf{X}\in\mathbb{R}^{(W-r+1)(H-r+1)\times Cr^2}$, $\mathbf{Y}\in\mathbb{R}^{(W-r+1)(H-r+1)\times P}$, and $\mathbf{F}\in\mathbb{R}^{Cr^2\times P}$.

Recall that the slice of $\mathcal{F}(\cdot,\cdot,i,j)$ is a circulant matrix.
Then according to the reshaping principle between $\mathcal{F}$ and $\mathbf{F}$, we have:
\begin{equation}
f_{a+C(i-1)+Cr(j-1),b}=f_{C(i-1)+Cr(j-1),b-a}, \forall a, b
\end{equation}
which means $\mathbf{F}$ is actually a block-circulant matrix. Hence the fast multiplication approach for block circulant matrix, as the ``FFT$\rightarrow$ component-wise multiplication $\rightarrow$IFFT" procedure, can now be applied to accelerate $\mathbf{Y}=\mathbf{XF}$, thereby resulting in the acceleration of (6). With the use of the proposed approach, the computational complexity for (6) is reduced from O($WHr^2CP$) to O($WHQ\log Q$), where $Q=\max(r^2C,P)$.

\subsection{Outline of Theoretical Proof}
\label{proof}
With the substantial reduction of weight storage and computational complexities, we attempt to prove that the proposed block-circulant matrix-based framework will consistently yield the similar overall accuracy compared with DNNs without compression. Only testing on existing benchmarks is insufficient given the rapid emergence of new application domains, DNN models, and data-sets. The theoretical proof will make the proposed method theoretically rigorous and distinct from prior work.

In the theory of neural networks, the ``\emph{effectiveness}" is defined using the \emph{universal approximation property}, which states that a neural network should be able to approximate any continuous or measurable function with arbitrary accuracy provided that an enough large number of parameters are available. This property provides the theoretical guarantee of using neural networks to solve machine learning problems, since machine learning tasks can be formulated as finding a proper approximation of an unknown, high-dimensional function. 
Therefore, the goal is to prove {\em the universal approximation property of block circulant matrix-based neural networks}, and more generally, for arbitrary structured matrices satisfying the low displacement rank $\gamma$. The detailed proofs for the block circulant matrix-based networks and general structured matrix-based ones are 
provided in the technical reports \cite{proof_simple,proof}.

The proof of the universal approximation property for block circulant matrix-based neural networks is briefly outlined as follows: Our objective is to prove that any continuous or measurable function can be approximated with arbitrary accuracy using a block-circulant matrix-based network. Equivalently, we aim to prove that the \emph{function space} achieved by block-circulant matrix-based neural networks is \emph{\textbf{dense}} in the space of continuous or measurable functions with the same inputs. An important property of the activation function, i.e., the \emph{component-wise discriminatory} property, is proved. Based on this property, the above objective is proved using proof by contradiction and Hahn-Banach Theorem \cite{narici1997hahn}.

We have further derived \emph{an approximation error bound of} O($1/n$) when the number of neurons in the layer $n$ is limited, with details shown in \cite{proof}. It implies that the approximation error will reduce with an increasing $n$, i.e., an increasing number of neurons/inputs in the network. As a result, we can guarantee the universal ``effectiveness" of the proposed framework on different DNN types and sizes, application domains, and hardware/software platforms.

\subsection{Compression Ratio and Test Accuracy}

\begin{figure*}[htbp]
\begin{center}
\includegraphics[width = 1\textwidth]{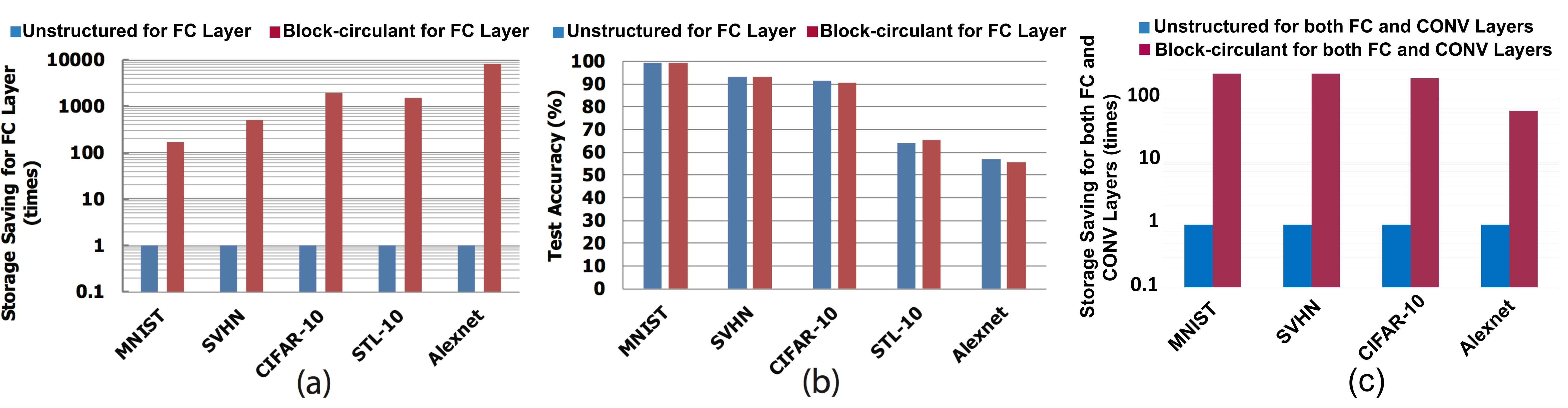}
\caption{(a) Storage saving and (b) test accuracy after using block-circulant FC layer for DCNN models on different datasets. (c) Storage saving after using both block-circulant FC layer and block-circulant CONV layer for DCNNs on MNIST, SVHN, CIFAR-10, and ImageNet datasets.}
\label{fig_SoftwareResult}
\end{center}
\vspace{-0.3em}
\end{figure*}

In this section, we apply \projectname to different DNN models 
{\em in software} and investigate the weight compression ratio and 
accuracy. 
Fig. \ref{fig_SoftwareResult} (a) and (b) show the weight storage (model size) reduction in FC layer and test accuracy on various image recognition datasets and DCNN models: MNIST (LeNet-5), CIFAR-10, SVHN, STL-10, and ImageNet (using AlexNet structure) \cite{krizhevsky2012imagenet,netzer2011reading,krizhevsky2009learning,coates2010analysis,deng2012mnist}). Here, 16-bit weight quantization is adopted for model size reduction. The baselines are the original DCNN models with unstructured weight matrices using 32-bit floating point representations. We see that block-circulant weight matrices enable 400$\times$-4000+$\times$ reduction in weight storage (model size) in corresponding FC layers. This parameter reduction in FC layers is also observed in \cite{cheng2015exploration}. The entire DCNN model size (excluding softmax layer) is reduced by 30-50$\times$ when only applying block-circulant matrices to the FC layer (and quantization to the overall network).
Regarding accuracy, the loss is negligible and sometimes 
the compressed models even outperform the baseline models. 

Fig. \ref{fig_SoftwareResult} (c) illustrates the further application of block-circulant weight matrices to the CONV layers on MNIST (LeNet-5), SVHN, CIFAR-10, and ImageNet (AlexNet structure) datasets, when {\em the accuracy degradation is constrained to be 1-2\%} by optimizing the block size. Again 16-bit weight quantization is adopted, and softmax layer is excluded. The 16-bit quantization also contributes to 2$\times$ reduction in model size. In comparison, the reductions of the number of parameters in \cite{han2015deep,han2015learning} are 12$\times$ for LeNet-5 (on MNIST dataset) and 9$\times$ for AlexNet. 
Moreover, another crucial property of \projectname is that
{\em the parameter storage after compression is regular}, whereas \cite{han2015deep,han2015learning} result in irregular weight storage patterns. The irregularity requires additional index per weight and significantly impacts the available parallelism degree.
From the results, we clearly see the significant benefit and 
potential of \projectname: it could produce highly compressed
models with regular structure.
\projectname yields more reductions in parameters compared with the state-of-the-art results for LeNet-5 and AlexNet. 
In fact, the actual gain could even be higher due to the indexing requirements of \cite{han2015deep,han2015learning}.  

We have also performed testing on other DNN models such as DBN, and found that \projectname can achieve similar or even higher compression ratio, demonstrating the wide application of block-circulant matrices. Moreover, a 5$\times$ to 9$\times$ acceleration in training can be observed for DBNs, which is less phenomenal than the model reduction ratio. This is because GPUs are less optimized for FFT operation than matrix-vector multiplications.

\section{{\large \projectname} Architecture}

Based on block-circulant matrix-based algorithms,
we propose \projectname architecture, --- a universal DNN inference engine that can be implemented in various hardware/software platforms with configurable network architecture (e.g., layer type, size, scales, etc.).

Applying \projectname to neural network accelerators enables notable architectural innovations. 
{\em 1)} Due to its recursive property and its intrinsic role in 
\projectname, FFT is implemented as the {\em basic computing block} (Section~\ref{fft_block}).
It ensures universal and small-footprint implementations.
{\em 2)} {\em Pipelining and parallelism} optimizations 
(Section~\ref{pipeline}). 
Taking advantage of the compressed but regular network structures, 
we aggressively apply inter-level and intra-level pipelining
in the basic computing block.
Moreover, we can conduct joint-optimizations considering 
parallelization degree, performance and power consumption.  
{\em 3)} {\em Platform-specific optimizations} focusing on 
weight storage and memory management.(Section~\ref{platform}).

\subsection{Recursive Property of FFT: the Key to Universal and 
Small Footprint Design}
\label{fft_block}

\begin{figure}[h]
\begin{center}
\includegraphics[width = 0.33\textwidth]{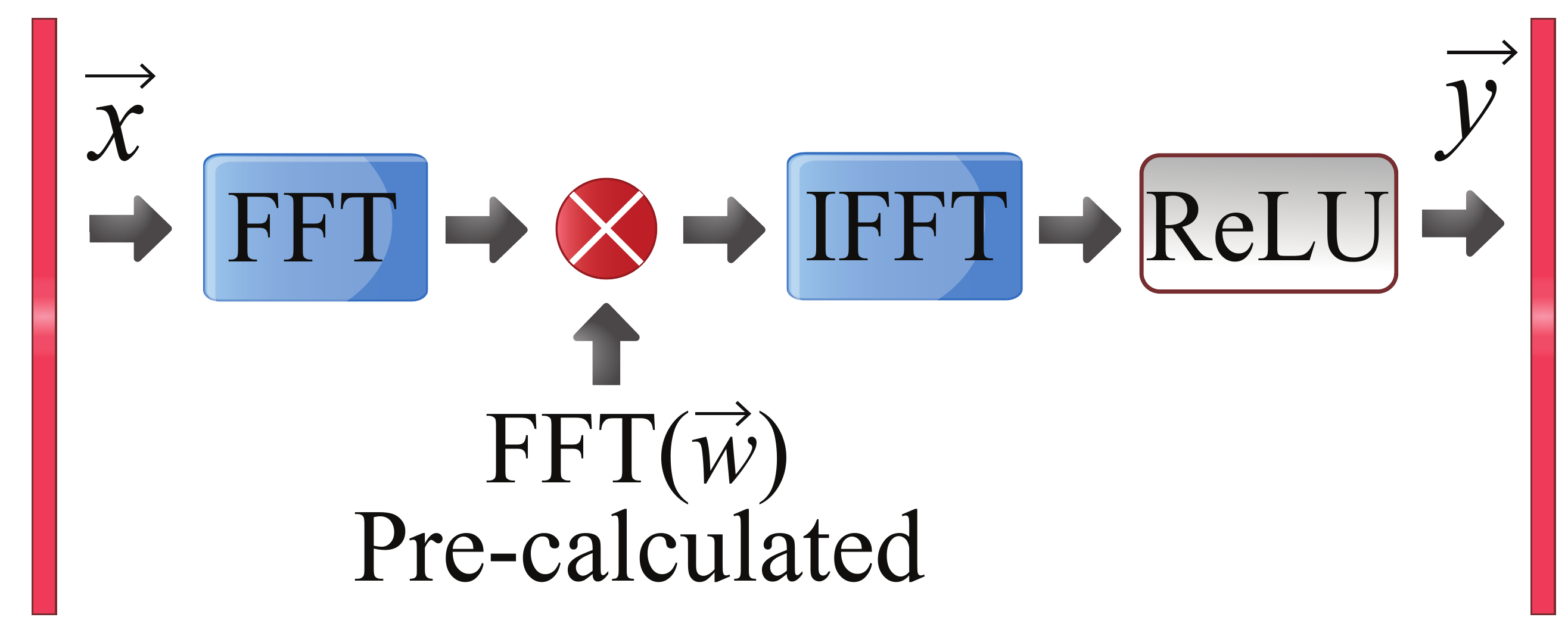}
\caption{The ``FFT$\rightarrow$component-wise multiplication$\rightarrow$IFFT" procedure.}
\label{fig_FFT}
\end{center}
\vspace{-0.3em}
\end{figure}

\begin{figure}[htbp]
\begin{center}
\includegraphics[width = 0.37\textwidth]{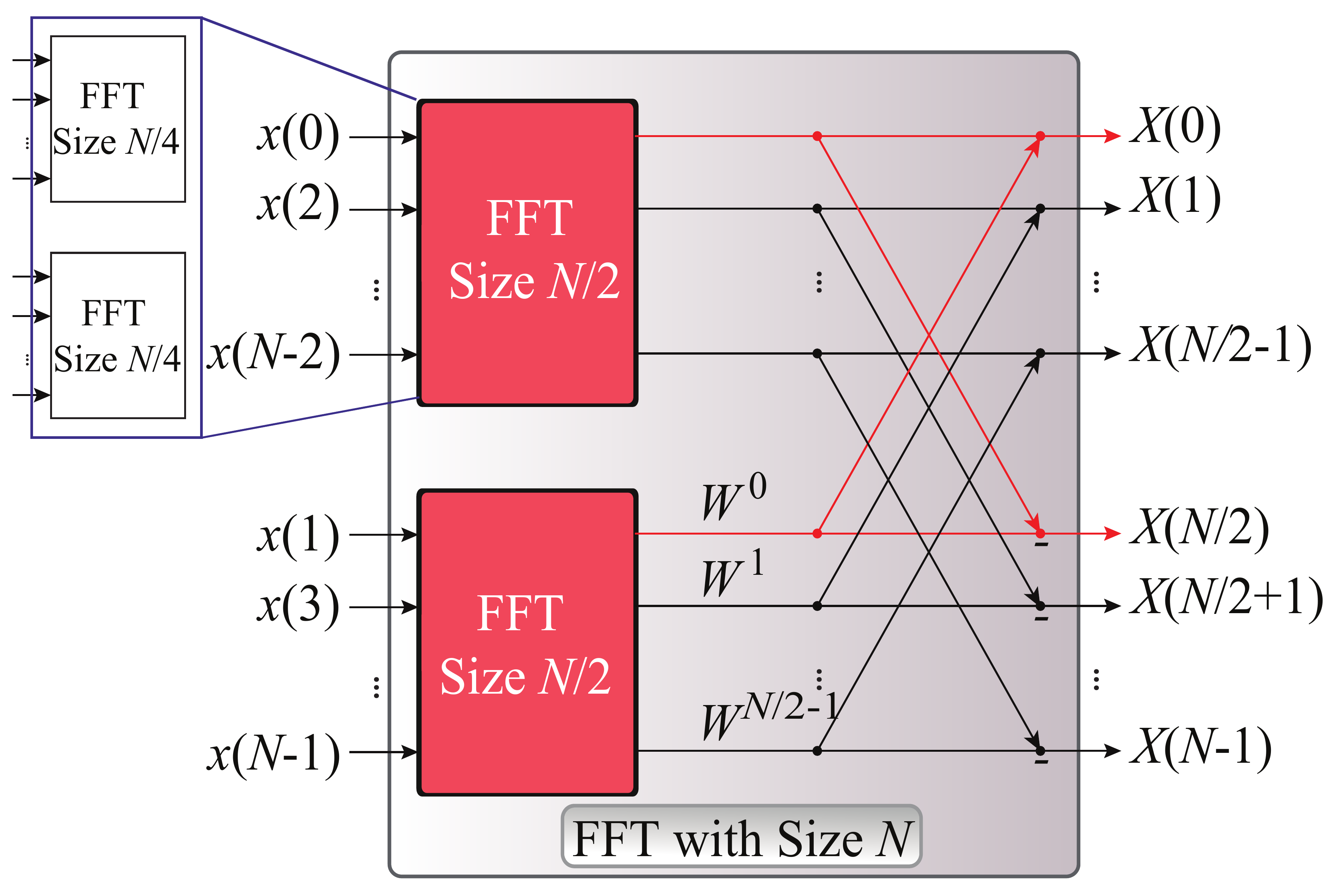}
\caption{Illustration of the recursive property of FFT.}
\label{fig_FFTRecursive}
\end{center}
\vspace{-0.3em}
\end{figure}

In \projectname, the ``FFT$\rightarrow$component-wise multiplication $\rightarrow$IFFT" in Fig. \ref{fig_FFT} 
is a universal procedure used in
both FC and CONV layers, for both inference and training processes, 
and for different DNN models.
We consider 
{\em FFT as the key computing kernel} in 
\projectname architecture due to its {\bf recursive property}.
It is known that FFT can be highly efficient with O($n\log n$) computational complexity, and hardware implementation of FFT has been investigated in  \cite{salehi2013pipelined,chang2003efficient,cheng2007high,garrido2009pipelined,garrido2009pipelined,ayinala2013place,ayinala2013fft}.
The recursive property states that 
the calculation of a size-$n$ FFT (with $n$ inputs and $n$ outputs) can be implemented using two FFTs with size $n/2$ plus one additional level of butterfly calculation, as shown in Fig. \ref{fig_FFTRecursive}. It can be further decomposed to four FFTs with size $n/4$ with two additional levels. 

The recursive property of FFT is the key to ensure
{\em a universal and reconfigurable} design which could handle 
different DNN types, sizes, scales, etc. 
It is because:
{\em 1)} A large-scale FFT can be calculated by recursively executing on the same computing block and some additional calculations; and
{\em 2)} IFFT can be implemented using the same structure as FFT with different preprocessing procedure and parameters \cite{salehi2013pipelined}.
It also ensures the design with small footprint, because:
{\em 1)} Multiple small-scale FFT blocks 
can be multiplexed and calculate a large-scale FFT with 
certain parallelism degree; and
{\em 2)} The additional component-wise multiplication has O($n$) complexity and relatively small hardware footprint.

\begin{figure}[htbp]
\begin{center}
\includegraphics[width = 0.45\textwidth]{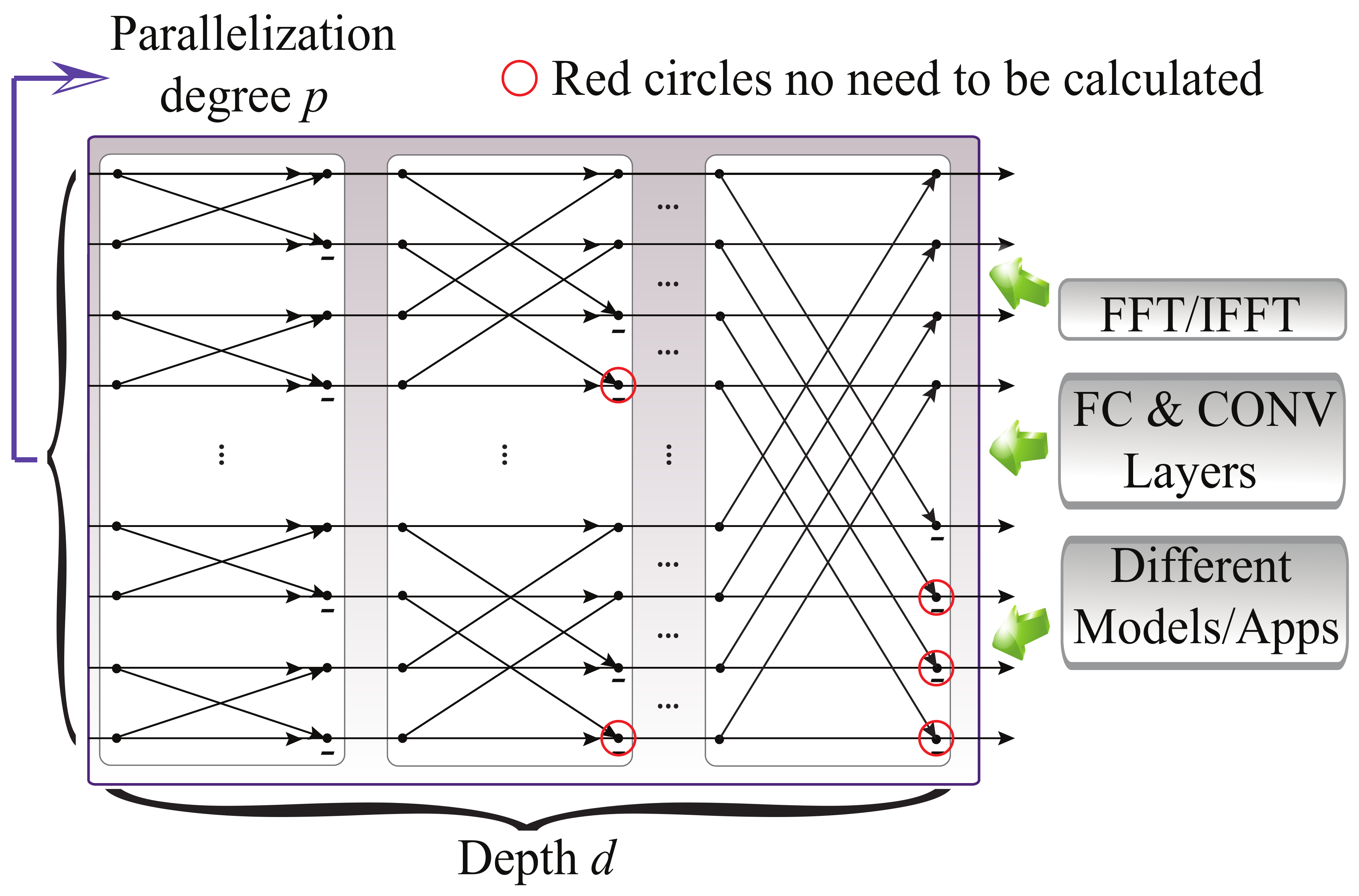}
\caption{The Basic Computing Block.}
\label{fig_Basic}
\end{center}
\vspace{-0.3em}
\end{figure}

Actual hardware systems, such as FPGA or ASIC designs, pose constraints on parallel implementation due to hardware footprint and logic block/interconnect resource limitations. As a result, we define the \emph{basic computing block} with a parallelization degree $p$ and depth $d$ (of butterfly computations), as shown in Fig. \ref{fig_Basic}. A \emph{butterfly computation} in FFT comprises cascade connection of complex number-based multiplications and additions \cite{oppenheim1999discrete,altera2010megacore}. The basic computing block is responsible for implementing the major computational tasks (FFT and IFFTs). An FFT operation (with reconfigurable size) is done by decomposition and iterative execution on the basic computing blocks.

Compared with conventional FFT calculation, we simplify the FFT computing based on the following observation: Our inputs of the deep learning system are from actual applications and are real values without imaginary parts. Therefore, the FFT result of each level will be a symmetric sequence except for the base component \cite{salehi2013pipelined}. As an example shown in the basic computing block shown in Fig. \ref{fig_Basic}, the partial FFT outcomes at each layer of butterfly computations will be symmetric, and therefore, the outcomes in the red circles do not need to be calculated and stored as partial outcomes. This observation can significantly reduce the amount of computations, storage of partial results, and memory traffic.

\subsection{Overall Architecture}
\label{overall}

\begin{figure}[htbp]
\begin{center}
\includegraphics[width = 0.35\textwidth]{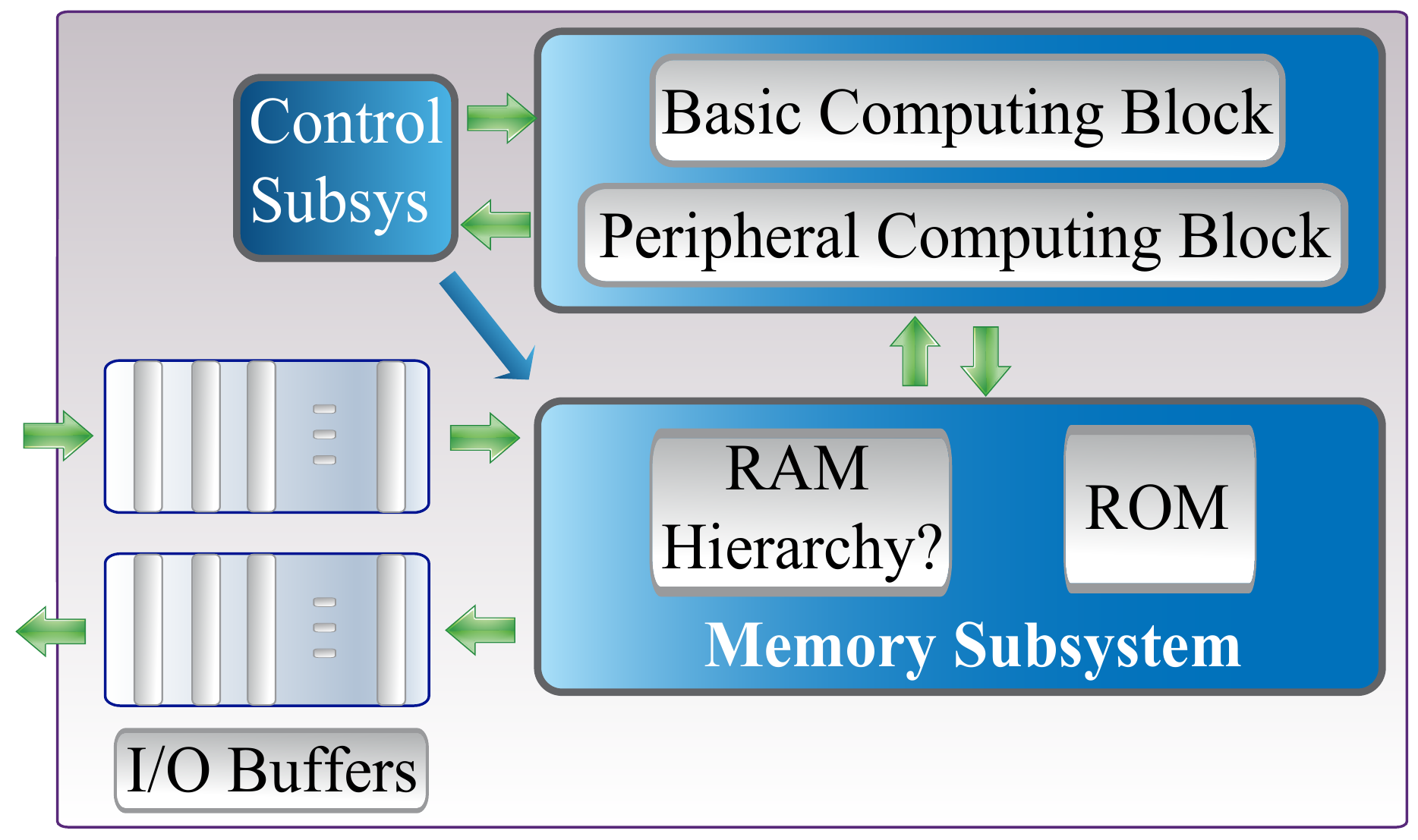}
\caption{\projectname Architecture.}
\label{fig_Overall}
\end{center}
\vspace{-0.3em}
\end{figure}

The overall \projectname architecture is shown in Fig. \ref{fig_Overall}, which includes the basic computing block, the peripheral computing block, the control subsystem, the memory subsystem, and I/O subsystem (I/O buffers). The basic computing block is responsible for the major FFT and IFFT computations. The peripheral computing block is responsible for performing component-wise multiplication, ReLU activation, pooling etc., which require lower (linear) computational complexity and hardware footprint. The implementations of ReLU activation and pooling are through comparators and have no inherent difference compared with prior work \cite{chen2014dadiannao,han2016eie}. The control subsystem orchestrates the actual FFT/IFFT calculations on the basic computing block and peripheral computing block. Due to the different sizes of CONV layer, FC layer and different types of deep learning applications, the 
different setting of FFT/IFFT calculations is 
configured by the control subsystem. The memory subsystem is composed of ROM, which is utilized to store the coefficients in FFT/IFFT calculations (i.e., the $W_n^i$ values including both real and imaginary parts); and RAM, which is used to store weights, e.g., the FFT results $\text{FFT}(\mathbf{w}_{ij})$. For ASIC design, a memory hierarchy may be utilized and carefully designed to ensure good performance.

We use 16-bit fixed point numbers for input and weight representations, which is common and widely accepted to be enough accurate for DNNs \cite{chen2014dadiannao,han2016eie,chen2017eyeriss,judd2016stripes}. Furthermore, it is pointed out \cite{han2015deep,lin2016fixed} that inaccuracy caused by quantization is largely independent of inaccuracy caused by compression and the quantization inaccuracy will not accumulate significantly for deep layers. 
%Hence 16-bit fixed point is safe for our framework both from theoretical analysis and actual experiments.

\subsection{Pipelining and Parallelism}
\label{pipeline}

\begin{figure}[htbp]
\begin{center}
\includegraphics[width = 0.48\textwidth]{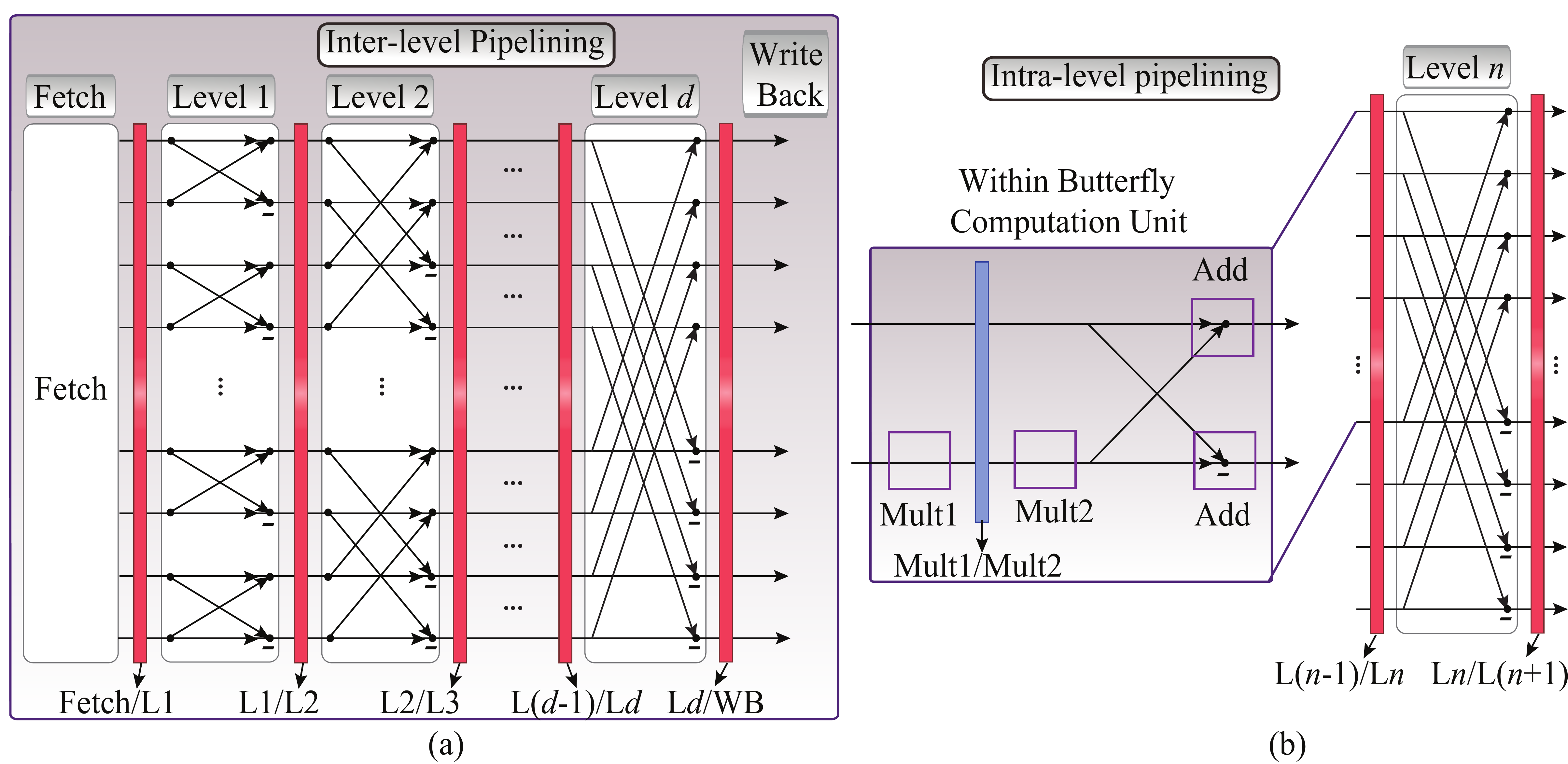}
\caption{The (a) inter-level and (b) intra-level pipelining techniques.}
\label{fig_Pipeline}
\end{center}
\vspace{-0.3em}
\end{figure}

Thanks to the regular block-circulant matrix structure, 
effective pipelining can be utilized 
to achieve the optimal tradeoff between energy efficiency and performance (throughput). 
\projectname architecture considers two pipelining techniques
as shown in Fig. \ref{fig_Pipeline}. 
In \emph{inter-level pipelining}, each pipeline stage corresponds to one level in the basic computing block. 
In \emph{intra-level pipelining}, 
additional pipeline stage(s) will be added within each butterfly computation unit, i.e., by deriving the optimal stage division in the cascade connection of complex number-based multiplication and additions. The proper selection of pipelining scheme highly depends on the target operating frequency and memory subsystem organization. In the experimental prototype, we target at a clock frequency around 200MHz (close to state-of-the-art ASIC tapeouts of DCNNs \cite{chen2017eyeriss,desoli201714,moons201714}), and therefore the  inter-level pipelining with a simpler structure will be sufficient for efficient implementations.

Based on the definition of $p$ (parallelization degree)
and $d$ (parallelization depth),
larger $p$ and $d$ values indicate higher level 
of parallelism and therefore will lead to higher performance and throughput, but also with higher hardware cost/footprint. A larger $d$ value would also result in less memory accesses at the cost of higher control complexity. We derive the optimal $p$ and $d$ values by
optimizing an overall metric $M\big(Perf(p,d),$ $Power(p,d)\big)$ as a suitable function of (average) performance and power consumption. The performance $Perf(p,d)$ will be an increasing function of $p$ and $d$ but also depends on the platform specifications, and the target type and size of DNN models (averaged over a set of learning models). The power consumption $Power(p,d)$ is a close-to-linear function of $pd$ accounting for both static and dynamic components of power dissipation. The optimization of $p$ and $d$ is constrained by the compute and memory hardware resource limits as well as memory and I/O bandwidth constraints. The proposed algorithm for design optimization is illustrated in Algorithm~\ref{algo}, which sets $p$ as optimization priority in order not to increase control complexity. This algorithm depends on the accurate estimation of performance and power consumption at each design configuration. 

We provide an example of design optimization and effects assuming a block size of 128 for FPGA-based implementation (Cyclone V). Because of the low operating frequency, increasing the $p$ value from 16 to 32 while maintaining $d=1$ only increases power consumption by less than 10\%. However, the performance can be increased by 53.8\% with a simple pipelining control. Increasing $d$ from 1 to 2 results in even less increase in power of 7.8\%, with performance increase of 62.2\%. The results seem to show that increasing $d$ is slightly more beneficial because of the reduced memory access overheads. However, a $d$ value higher than 3 will result in high control difficulty and pipelining bubbles, whereas $p$ can be increased with the same control complexity thanks to the high bandwidth of block memory in FPGAs. As a result, we put $p$ as the optimization priority in Algorithm 3.

% \begin{figure}[htbp]
% \begin{center}
% \includegraphics[width = 0.1\textwidth]{Figures/Fig_Overall1.pdf}
% \captionsetup{labelformat=empty}
% \caption{}
% \end{center}
% \end{figure}

\begin{algorithm}[t]\small			

	\emph{Optimize parallel degree $p$}:\\
	\quad Derive upper bound of $p$ based on memory bandwidth \\
    \qquad -limit \& hardware resource limit;\\
    \quad Use ternary search for $p$:\\ 	 \qquad Estimate $M\big(Perf(p,d),$ $Power(p,d)\big)$  assuming $d=1$;\\Optimize depth $d$ using the ternary search method, based on the derived $p$ value. 
	
\caption{\footnotesize The proposed  algorithm for design optimization of basic computing block}\label{algo}
\end{algorithm}

\subsection{Platform-Specific Optimizations}
\label{platform}

Based on the generic \projectname architecture, this section 
describes platform-specific optimizations on 
the FPGA-based and ASIC-based hardware platforms.
We focus on weight storage and memory management, in order to simplify the design and achieve higher energy efficiency and performance.

\noindent
{\bf FPGA Platform}.
The {\em key observation} is that the weight storage requirement of representative DNN applications can be (potentially) met by the on-chip block memory in state-of-the-art FPGAs. As a representative large-scale DCNN model for ImageNet application, the whole AlexNet \cite{krizhevsky2012imagenet} results in only {\em around 4MB storage} requirement after (i) applying block-circulant matrices only to FC layers, and (ii) using 16-bit fixed point numbers that results in negligible accuracy loss. Such storage requirement can be fulfilled by the on-chip block memory of state-of-the-art FPGAs such as Intel (former Altera) Stratix, Xilinx Virtex-7, etc., which consist of up to tens of MBs on-chip memory \cite{link1,link2}. 
Moreover, when applying block-circulant matrices also to CONV layers,
the storage requirement can be further 
{\em reduced to 2MB or even less} (depending on the block size and tolerable accuracy degradation).
%In addition, the storage requirement is further {\em reduced to from hundreds of KBs to 2MB} when applying block-circulant matrices to CONV layers (depending on the block size) and performing compression on FFT results stored in memory. 
Then, the storage requirement becomes comparable with the input size and can be potentially supported by low-power and high energy-efficiency FPGAs such as Intel (Altera) Cyclone V or Xilinx Kintex-7 FPGAs. A similar observation holds for other applications (e.g., the MNIST data-set \cite{deng2012mnist}), or different network models like DBN or RNN. Even for future larger-scale applications, the full model after compression will (likely) fit in an FPGA SoC leveraging the storage space of the integrated ARM core and DDR memory \cite{link1,link2}. This observation would make the FPGA-based design significantly more efficient.

In state-of-the-art FPGAs, on-chip memory is organized in the form of memory blocks, each with certain capacity and bandwidth limit. The number of on-chip memory blocks represents a proper tradeoff between the lower control complexity (with more memory blocks) and the higher energy efficiency (with fewer memory blocks), and thus should become an additional knob for design optimizations. State-of-the-art FPGAs are equipped with comprehensive DSP resources such as $18\times 18$ or variable-size multipliers \cite{link1,link2}, which are effectively exploited for performance and energy efficiency improvements.

\noindent
{\bf ASIC platform}.
We mainly investigate two aspects in the memory subsystem: 
{\em 1)} the potential memory hierarchy and 
{\em 2)} the memory bandwidth and aspect ratio. The representative deep learning applications require hundreds of KBs to multiple MBs memory storage depending on different compression levels, and we assume a conservative value of multiple MBs due to the universal and reconfigurable property of \projectname architecture. 
The potential memory hierarchy structure depends strongly on the target clock frequency of the proposed system. Specifically, if we target at a clock frequency around 200MHz (close to state-of-the-art ASIC tapeouts of DCNNs \cite{chen2017eyeriss,desoli201714,moons201714}), then the memory hierarchy is not necessary because a single-level memory system can support such operating frequency. Rather, memory/cache reconfiguration techniques \cite{wang2011dynamic,gordon2008phase} can be employed when executing different types and sizes of applications for performance enhancement and static power reduction. If we target at a higher clock frequency, say 800MHz, an effective memory hierarchy with at least two levels (L1 cache and main memory) becomes necessary because a single-level memory cannot accommodate such high operating frequency in this case. Please note that the cache-based memory hierarchy is highly efficient and results in very low cache miss rate because, \emph{prefetching} \cite{jouppi1998improving,dundas1997improving}, the key technique to improve performance, will be highly effective due to the regular weight access patterns. 
The effectiveness of prefetching is due to 
%Such memory access optimization inherently benefits from 
the {\em regularity} in the proposed block-circulant matrix-based neural networks, showing another advantage over prior compression schemes. In our experimental results in the next section, we target at a lower clock frequency of 200MHz and therefore the memory hierarchy structure is not needed.

Besides the memory hierarchy structure, the memory bandwidth is determined by the parallelization degree $p$ in the basic computing block. Based on such configuration, the aspect ratio of the memory subsystem is determined. Because of the relatively high memory bandwidth requirement compared with the total memory capacity (after compression using block-circulant matrices), column decoders can be eliminated in general~\cite{weste2005cmos}, thereby resulting in simpler layout and lower routing requirements.

\section{Evaluation}
\label{eval}

In this section, we provide detailed experimental setups and results of the proposed universal inference framework on different platforms including FPGAs, ASIC designs, and embedded processors.
Experimental results on representative benchmarks such as MNIST, CIFAR-10, SVHN, and ImageNet have been provided and we have conducted a comprehensive comparison with state-of-the-art works on hardware deep learning systems. Order(s) of magnitude in energy efficiency and performance improvements can be observed using the proposed universal inference framework.

\subsection{FPGA-Based Testing}
First, we illustrate our FPGA-based testing results using a low-power and low-cost Intel (Altera) Cyclone V 5CEA9 FPGA. The Cyclone V FPGA exhibits a low static power consumption less than 0.35W and highest operating frequency between 227MHz and 250MHz (but actual implementations typically have less than 100MHz frequency),
making it 
%and can be utilized 
a good choice for energy efficiency optimization of FPGA-based deep learning systems.
\begin{figure}[htbp]
\begin{center}
\includegraphics[width = 0.37\textwidth]{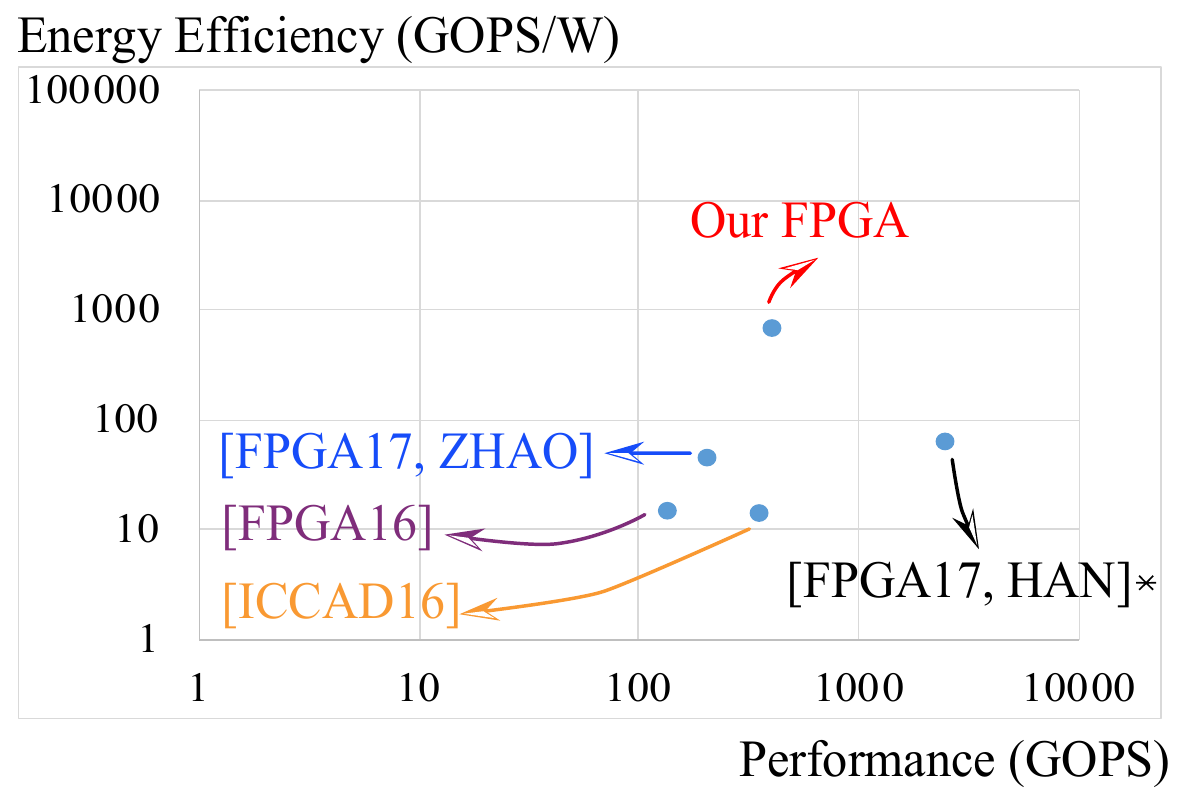}
\caption{Comparison on performance and energy efficiency with state-of-the-art FPGA results.}
\label{fig_fpga}
\end{center}
\vspace{-0.3em}
\end{figure}

Fig.~\ref{fig_fpga} illustrates the comparison of performance (in giga operations per second, GOPS) and energy efficiency (in giga operations per Joule, GOPS/W) between the proposed and reference FPGA-based implementations. The FPGA implementation uses the AlexNet structure, a representative DCNN model with five CONV layers and three FC layers for the ImageNet applications \cite{deng2009imagenet}.
The reference FPGA-based implementations are state-of-the-arts represented by [FPGA16] \cite{qiu2016going}, [ICCAD16] \cite{zhang2016caffeine}, [FPGA17, Han] \cite{han2017ese}, and [FPGA17, Zhao] \cite{zhao2017accelerating}. The reference works implement large-scale AlexNet, VGG-16, medium-scale DNN for CIFAR-10, or a custom-designed recurrent neural network \cite{han2017ese}. Note that we use equivalent GOPS and GOPS/W for all methods with weight storage compression, including ours. Although those references focus on different DNN models and structures, both GOPS and GOPS/W are general metrics that are independent of model differences. It is widely accepted in the hardware deep learning research to compare the GOPS and GOPS/W metrics between their proposed designs and those reported in the reference work, as shown in \cite{qiu2016going,zhang2016caffeine,zhao2017accelerating}. Please note that this is not entirely fair comparison because our implementation is layerwise implementation (some reference works implement end-to-end networks) and we extracts on-chip FPGA power consumptions.

In Fig.~\ref{fig_fpga}, we can observe the significant improvement achieved by the proposed FPGA-based implementations compared with prior arts in terms of energy efficiency, even achieving 11$\times$-16$\times$ improvement when comparing with prior work with heuristic model size reduction techniques \cite{han2017ese,zhao2017accelerating} (reference \cite{han2017ese} uses the heuristic weight pruning method, and \cite{zhao2017accelerating} uses a binary-weighted neural network XOR-Net). When comparing with prior arts with a uncompressed (or partially compressed) deep learning system \cite{qiu2016going,zhang2016caffeine}, the energy efficiency improvement can reach 60-70$\times$. These results demonstrate a clear advantage of 
\projectname using block-circulant matrices on energy efficiency.
The performance and energy efficiency improvements are due to:
{\em 1)} algorithm complexity reduction and {\em 2)} efficient hardware design, weight reduction, and elimination of weight accessing to the off-chip storage. The first source results in 10$\times$-20$\times$ improvement and the second results in 2$\times$-5$\times$.
Please note that \projectname architecture 
does not yield the highest throughput because we use a single low-power FPGA, while reference \cite{han2017ese} uses a high-performance FPGA together with large off-chip DRAM on a custom-designed recurrent neural network. If needed, we can increase the number of FPGAs to process multiple neural networks in parallel, thereby improving the throughput without incurring any degradation in the energy efficiency.

Besides comparison with large-scale DCNNs with
\projectname architecture with state-of-the-art FPGA implementations, we also compare it with IBM TrueNorth neurosynaptic processor on a set of benchmark data sets including MNIST, CIFAR-10, and SVHN, which are all supported by IBM TrueNorth\footnote{Please note that AlexNet is not currently supported by IBM TrueNorth due to the high-degree neural connections.}. This is a fair comparison because both the proposed system and IBM TrueNorth are end-to-end implementations. IBM TrueNorth \cite{esser2016convolutional} is a neuromorphic CMOS chip fabricated in 28nm technology, with 4096 cores each simulating 256 programmable silicon neurons in a time-multiplexed manner. It implements \emph{Spiking Neural Networks}, which is a bio-inspired type of neural networks and benefits from the ability of globally asynchronous implementations, but is widely perceived to achieve a lower accuracy compared with state-of-the-art DNN models. IBM TrueNorth exhibits the advantages of reconfigurability and programmability. Nevertheless, reconfigurability also applies to 
\projectname architecture.

\begin{figure}[htbp]
\begin{center}
\includegraphics[width = 0.48\textwidth]{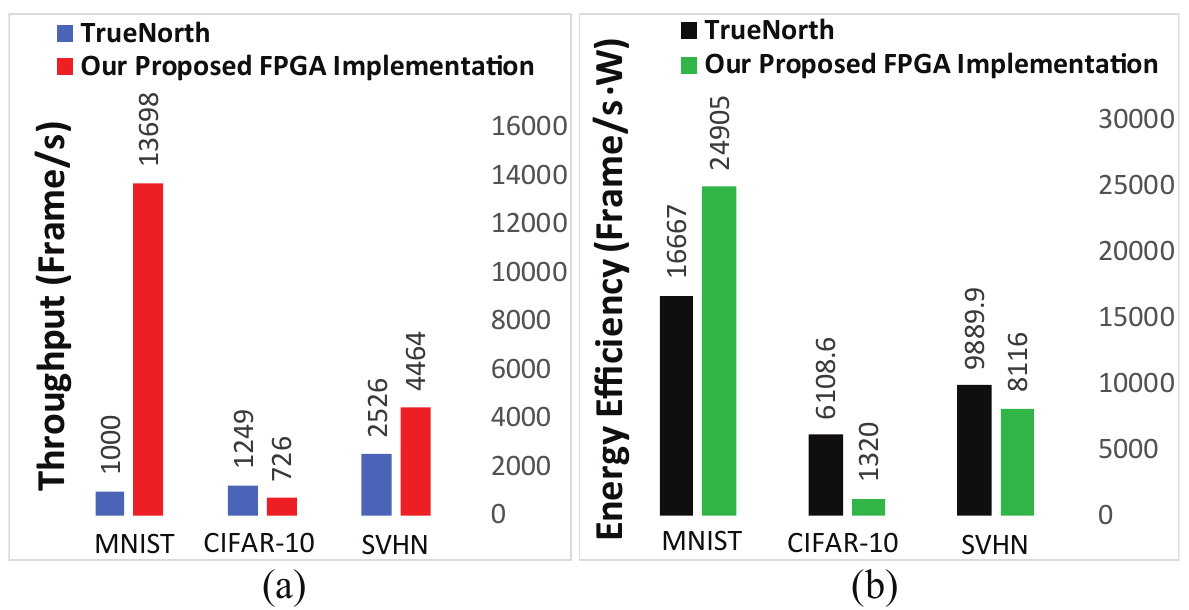}
\caption{Comparison on (a) throughput and (b) energy efficiency between the proposed FPGA end-to-end implementation with TrueNorth reports~\cite{esser2015backpropagation,esser2016convolutional}.}
\label{fig_fpga_1}
\end{center}
\vspace{-0.3em}
\end{figure}

Fig. \ref{fig_fpga_1} compares the throughput and energy efficiency of FPGA-based implementations of \projectname architecture and IBM TrueNorth on different benchmark data sets. The throughput and energy efficiency of IBM TrueNorth results are from~\cite{esser2015backpropagation,esser2016convolutional} (\cite{esser2015backpropagation} for MNIST and \cite{esser2016convolutional} for CIFAR-10 and SVHN), and we choose results from the low-power mapping mode using only a single TrueNorth chip for high energy efficiency. We can observe the improvement on the throughput for MNIST and SVHN data sets and energy efficiency on the same level of magnitude. The throughput of CIFAR-10 using the FPGA implementation of \projectname is lower because {\em 1)} TrueNorth requires specific preprocessing of CIFAR-10 ~\cite{esser2016convolutional} before performing inference, and 
{\em 2)} the DNN model we chose uses small-scale FFTs, which limits the degree of improvements. Besides, \projectname architecture achieves the higher test accuracy in general: it results in {\em very minor accuracy degradation compared with software DCNNs} (cf. Fig. \ref{fig_SoftwareResult}), whereas the low-power mode of IBM TrueNorth incurs higher accuracy degradation~\cite{esser2015backpropagation,esser2016convolutional}. These results demonstrate the high effectiveness of \projectname architecture
because it is widely perceived that FPGA-based implementations will result in lower performance and energy efficiency compared with ASIC implementations, with benefits of a short development round and higher flexibility.

\subsection{ASIC Designs Synthesis Results}

\begin{figure}[htbp]
\begin{center}
\includegraphics[width = 0.37\textwidth]{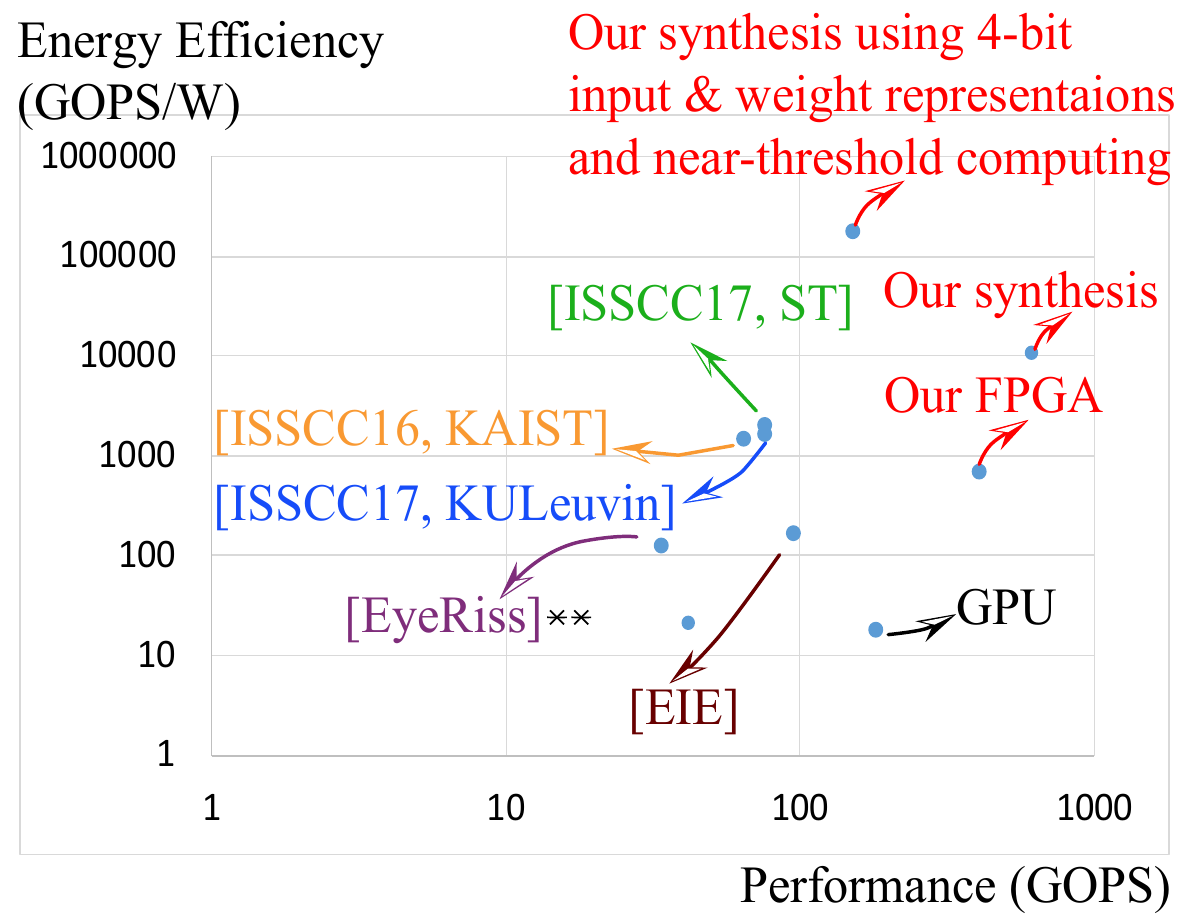}
\caption{Comparison on performance and energy efficiency with state-of-the-art ASIC results.}
\label{fig_ASIC}
\end{center}
\vspace{-0.3em}
\end{figure}

We derive ASIC synthesis results of \projectname architecture
for large-scale DCNN implementations (AlexNet) and compare with state-of-the-art ASIC developments and synthesis results. The delay, power, and energy of our ASIC designs are obtained from synthesized RTL under Nangate 45nm process \cite{nangate} using Synopsys Design Compiler. The memories are SRAM based and estimated using CACTI 5.3 \cite{cacti}. Fig.~\ref{fig_ASIC} illustrates the comparison results on ASIC-based implementations/synthesis with state-of-the-arts, which also target at large-scale deep learning systems. The reference ASIC implementations and synthesis results are represented by [EIE] \cite{han2016eie}, [Eyeriss] \cite{chen2017eyeriss}, [ISSCC16, KAIST] \cite{sim20161}, [ISSCC17, ST] \cite{desoli201714}, and [ISSCC17, KULeuvin] \cite{moons201714}. 

In Fig.~\ref{fig_ASIC}, we can observe that our synthesis results achieve both the highest throughput and energy efficiency, more than 6 times compared with the highest energy efficiency in the best state-of-the-art implementations. It is also striking that even our FPGA implementation could achieve the same order of energy efficiency and higher throughput compared with the best state-of-the-art ASICs. It is worth noting that the best state-of-the-art ASIC implementations report the highest energy efficiency in the near-threshold regime with an aggressively reduced bit-length (say 4 bits, with a significant accuracy reduction in this case). When using 4-bit input and weight representations and near-threshold computing of 0.55V $V_{dd}$ voltage level, another 17$\times$ improvement on energy efficiency can be achieved in the synthesis results compared with our super-threshold implementation, as shown in Fig.~\ref{fig_ASIC}. This makes it a total of 102$\times$ improvement compared with the best state-of-the-art. Moreover, in our systems (in the super-threshold implementation), memory in fact consumes slightly less power consumption compared with computing blocks, which demonstrates that weight storage is no longer the system bottleneck. 
Please note that the overall accuracy when using 4-bit representation is low in \projectname architecture (e.g., less than 20\% for AlexNet). Hence, 4-bit representation is only utilized to provide a fair comparison with the baseline methods using the same number of bits for representations. 

We also perform comparison on performance and energy efficiency with the most energy-efficient NVIDIA Jetson TX1 embedded GPU toolkit, which is optimized for deep learning applications. It can be observed that an energy efficiency improvement of 570$\times$ can be achieved using our implementation, and the improvement reaches 9,690$\times$ when incorporating near-threshold computing and 4-bit weight and input representations. 

\subsection{Embedded ARM-based Processors}
Because ARM-based embedded processors are the most widely used embedded processors in smartphones, embedded and IoT devices, we implement the proposed block-circulant matrix-based DNN inference framework on a smartphone using ARM Cortex A9 processor cores, and provide some sample results. The aim is to demonstrate the potential of real-time implementation of deep learning systems on embedded processors, thereby significantly enhancing the wide adoption of (large-scale) deep learning systems in personal, embedded, and IoT devices. In the implementation of LeNet-5 DCNN model on the MNIST data set, the proposed embedded processor-based implementation achieves a performance of 0.9ms/image with 96\% accuracy, which is slightly faster compared with IBM TrueNorth in the high-accuracy mode \cite{esser2015backpropagation} (1000 Images/s). The energy efficiency is slightly lower but at the same level due to the peripheral devices in a smartphone. When comparing with a GPU-based implementation using NVIDIA Tesla C2075 GPU with 2,333 Images/s, the energy efficiency is significantly higher because the GPU consumes 202.5W power consumption, while the embedded processor
only consumes around 1W. It is very interesting that when comparing on the fully-connected layer of AlexNet, our smartphone-based implementation of \projectname even achieves higher throughput (667 Layers/s vs. 573 Layers/s) compared with NVIDIA Tesla GPU. This is because the benefits of computational complexity reduction become more significant when the model size becomes larger.

\subsection{Summary and Discussions}
\textbf{\emph{Energy Efficiency and Performance:}} Overall, 
\projectname architecture achieves a significant gain in energy efficiency and performance compared with the best state-of-the-arts
on different platforms including FPGAs, ASIC designs, and embedded processors. The key reasons of such improvements include the fundamental algorithmic improvements, weight storage reduction, a significant reduction of off-chip DRAM accessing, and the highly efficient implementation of the basic computing block for FFT/IFFT calculations. The fundamental algorithmic improvements accounts for the most significant portion of energy efficiency and performance improvements around 10$\times$-20$\times$, and the rest accounts for 2$\times$-5$\times$.
In particular, we emphasize that: 
{\em 1)} the hardware resources and power/energy consumptions associated with memory storage will be at the same order as the computing blocks and will not be the absolute dominating factor of the overall hardware deep learning system; 
{\em 2)} medium to large-scale DNN models can be implemented in small footprint thanks to the recursive property of FFT/IFFT calculations. These characteristics are the key to enable highly efficient implementations of \projectname architecture in low-power FPGAs/ASICs and the elimination of complex control logics and high-power-consumption clock networks.

\textbf{\emph{Reconfigurability:}} 
It is a key property of \projectname architecture,
allowing it be applied to a wide set of deep learning systems.
It resembles IBM TrueNorth and could significantly reduce the development round and promote the wide application of deep learning systems. Unlike IBM TrueNorth, 
\projectname {\em 1)} does not need a specialized offline training framework and specific preprocessing procedures for certain data sets like CIFAR \cite{esser2016convolutional}; and 
{\em 2)} does not result in any hardware resource waste for small-scale neural networks and additional chips for large-scale ones. The former property is because the proposed training algorithms are general, and the latter is because different scales of DNN models can be conducted on the same basic computing block using different control signals thanks to the recursive property of FFT/IFFT. The software interface of reconfigurability is under development and will be released for public testing.

\textbf{\emph{Online Learning Capability:}} 
The \projectname architecture described mainly focuses on the inference process of deep learning systems, although its algorithmic framework applies to both inference and training. We focus on inference because it is difficult 
to perform online training in hardware embedded deep learning systems due to the limited computing power and data set they can encounter. %However, the recent development of multi-modal intelligent systems such as unmanned driving and unmanned aerial systems necessitate the ability of performing online adaption especially for the mission control and trajectory generation. 
%To accommodate these needs and account for the limitation of online data, we adopt the similar idea as ``transfer learning''~\cite{pan2010survey} of only updating the weights in the FC layers in the online learning procedure, and are currently developing the first (according to our knowledge) low-power hardware deep learning system incorporating the online learning property on a partially learnt DNN model. Effective back-propagation-like algorithm with complexity O$(n\log n)$ per layer and moderate storage complexity for minibatch updating, as shown in Section 4.1, will be adopted in the framework.  

\section{Conclusion}

This paper proposes \projectname,
a principled approach to represent weights and process neural networks using 
{\em block-circulant} matrices.
% for weight representation.
%We overcome these limitations by proposing block-circulant matrices for weight representation in deep learning systems.
\projectname utilizes the {\em Fast Fourier Transform 
(FFT)}-based fast multiplication, {\em simultaneously} reducing 
the computational complexity (both in inference and training) from O($n^2$) to O($n\log n$) and 
the storage complexity from O($n^2$) to O($n$), with negligible accuracy loss. 
We propose the \projectname architecture, a universal DNN inference engine that can be implemented in various hardware/software 
platforms with configurable network architecture (e.g., layer type,
size, scales, etc.).
To demonstrate the performance and energy efficiency,
we test \projectname architecture in FPGA, ASIC and embedded processors.
Our results show that \projectname architecture achieves
very high energy efficiency and 
performance with a small hardware footprint.
Based on the FPGA implementation and ASIC synthesis results, 
\projectname achieves 6 - 102$\times$ energy efficiency improvements compared with the best state-of-the-art results. 

\section{Acknowledgement}

This work is funded by the National Science Foundation Awards CNS-1739748, CNS-1704662, CNS-1337300, CRII-1657333, CCF-1717754, CNS-1717984, Algorithm-in-the-Field program, DARPA SAGA program, and CASE Center at Syracuse University.

%\bibliographystyle{ieeetr}
%\bibliographystyle{ACM-Reference-Format}
%\bibliography{ref} 

\end{document}